\RequirePackage{fix-cm}
\documentclass[smallextended]{svjour3}       
\smartqed  
\usepackage{graphicx}
\usepackage{amsfonts}
\usepackage{amsmath}
\usepackage{amssymb}
\usepackage{multicol}
\usepackage[table,xcdraw]{xcolor}
\usepackage{lscape}
\begin{document}

\title{Person Retrieval in Surveillance Using Textual Query: A Review
}



\author{Hiren Galiyawala         \and
	Mehul S. Raval 
}


\institute{Hiren Galiyawala \at
	School of Engineering and Applied Science, \\
	Ahmedabad University, India.\\
	\email{hirenkumar.g@ahduni.edu.in}           
	\and
	Mehul S. Raval \at
	School of Engineering and Applied Science, \\
	Ahmedabad University, India.\\
	\email{mehul.raval@ahduni.edu.in} 
}

\date{Received: date / Accepted: date}

\maketitle

\begin{abstract}

Recent advancement of research in biometrics, computer vision, and natural language processing has discovered opportunities for person retrieval from surveillance videos using textual query. The prime objective of a surveillance system is to locate a person using a description, e.g., \textit{a short woman with a pink t-shirt and white skirt carrying a black purse. She has brown hair}. Such a description contains attributes like gender, height, type of clothing, colour of clothing, hair colour, and accessories. Such attributes are formally known as \textit{soft biometrics}. They help bridge the semantic gap between a human description and a machine as a textual query contains the person's soft biometric attributes. It is also not feasible to manually search through huge volumes of surveillance footage to retrieve a specific person. Hence, automatic person retrieval using vision and language-based algorithms is becoming popular. In comparison to other state-of-the-art reviews, the contribution of the paper is as follows: 1. Recommends most discriminative soft biometrics for specific challenging conditions. 2. Integrates benchmark datasets and retrieval methods for objective performance evaluation. 3. A complete snapshot of techniques based on features, classifiers, number of soft biometric attributes, type of the deep neural networks, and performance measures. 4. The comprehensive coverage of person retrieval from handcrafted features based methods to end-to-end approaches based on natural language description.

\end{abstract}

\section{Introduction and motivation}
\label{sec:1}
Security of the individual and society is a significant concern in today's world of automation. Modern life of human beings involves many Internet-of-Things (IoT) based devices to interact and operate. While such devices provide luxury to live, they may also endanger living by compromising privacy. There are many questions in daily life such as:

\begin{itemize}
	\item Is this person authorized to enter into specific premises?
	\item Is this person authorized to access the mobile application?
	\item Is the right person accessing the system?
	\item Is the person a legal resident of the country?
\end{itemize}

Answers to such questions need monitoring using surveillance networks. Most countries in the world have surveillance cameras in many places. Camera networks cover many places like housing colonies, small offices, large corporate offices, retail shops, supermarkets, public places like airports, railway stations, and sports venues. The world has observed many events like terrorist attacks and general strikes in recent years. Public safety is the most critical task to be carried out by a security agency during such an event. A surveillance network helps investigators during and after the event. Person localisation plays a significant role during such an investigation and helps to accelerate the process of investigation. Typically, video operators carry out the person(s) retrieval task by searching through a video database. It takes time, and the process is also very inefficient. Thus, intelligent video surveillance is growing as an active research area.

Some critical definitions for giving the reader a better understanding are as follows:

\paragraph{Person identification:}
It is a process to establish an individual's identity or to identify an individual uniquely. Biometrics such as fingerprint and face are unique to an individual, which helps in person identification.

\paragraph{Person re-identification (Re-ID):}
Re-ID aims to identify a person in a new place and at a different time captured by another camera. The query can be in the form of an image, video, and even text description. It is useful to re-identify a person in surveillance-based applications.

\paragraph{Person retrieval}
aims to spot the person of interest from an image gallery or video frame given the query. It does not seek to establish an identity; instead, it reduces the search space. Thus, the output may contain multiple retrievals matching the query. For example, a surveillance frame may have multiple persons with a description, i.e., \textit{a man with a blue t-shirt and black jeans}.

\subsection{History and background}
\label{sec:1.1}

\begin{figure}
	\centering
	\includegraphics[width=\columnwidth]{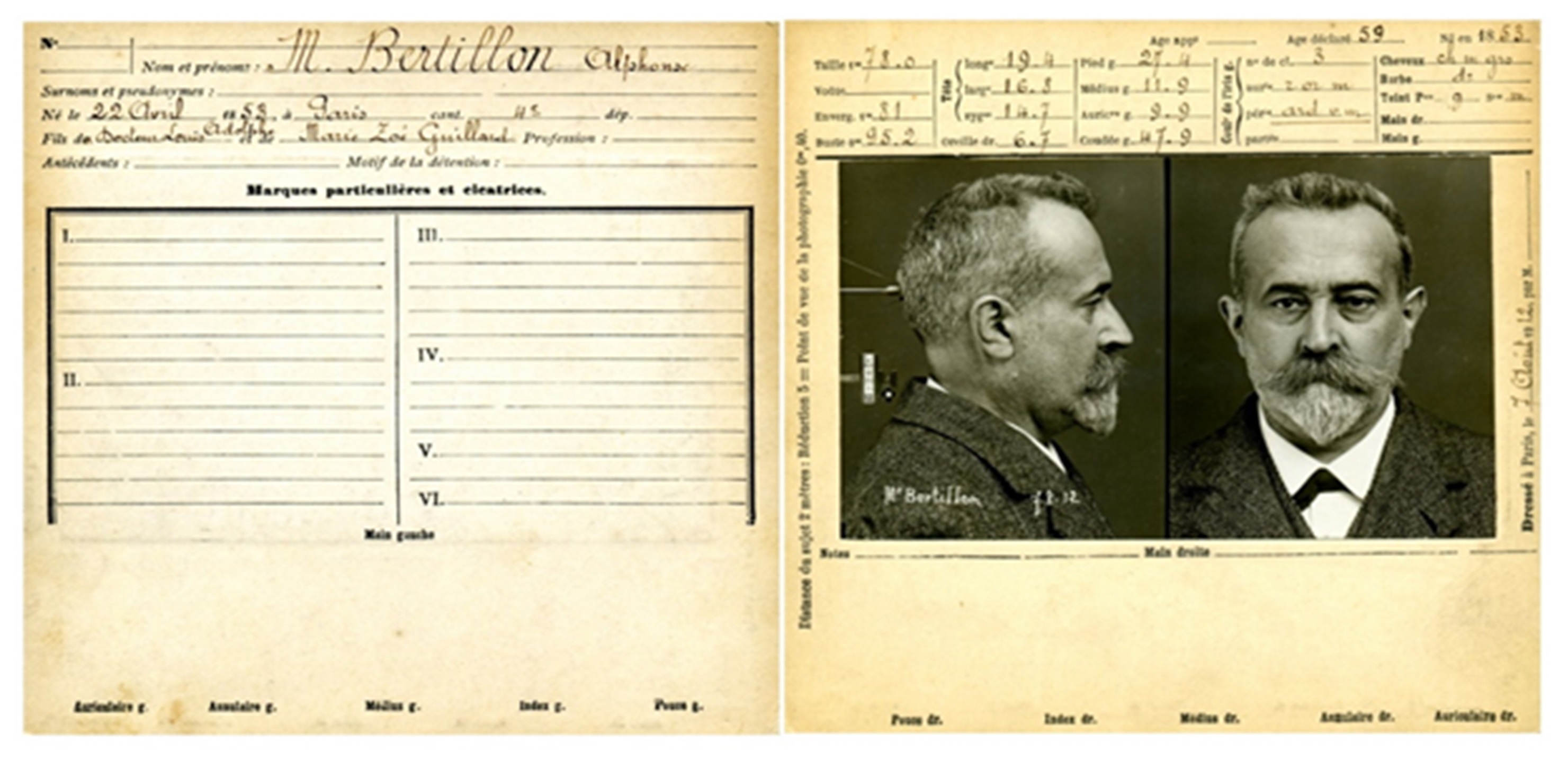}
	\caption{Anthropometric data sheet of Alphonse Bertillon \cite{R3}.}
	\label{fig:1}       
\end{figure}

An early person identification system developed by Alphonse Bertillon, a police officer and biometric researcher from Paris, France, identified criminals in the 19th century. His approach is also known as Bertillonage \cite{R1}, which was widely accepted in France and other countries. Criminal identification was made only by employing photographs or names before Bertillonage. He was the first to use anthropometry for law enforcement using various physical measurements. He also defined the process of physical measurements and photograph acquisition to maintain uniformity in the records \cite{R1,R2} (see Fig.~\ref{fig:1}). Bertillon's system measures physical features like standing height, length and breadth of head, length of individual fingers, dimensions of the foot, dimensions of the nose, and ear. It also has descriptions of eye and hair colour, any scars, marks, and tattoos on the body \cite{R1}.

Two persons may likely have the same height, but the chances of other measurements being similar is highly unlikely. Sizes are not unique to individuals and hence may fail to identify a person if used without other attributes. e.g., twins may have similar biological features. Hence, it has often been superseded by fingerprint-based identification. Biometrics is a reliable solution for person identification \cite{R4} due to properties like uniqueness, stability, universality, and accessibility. The answer is dependent on \textquotedblleft what you are\textquotedblright which uses face, fingerprint, palm print, hand geometry.  It is independent of \textquotedblleft what you possess\textquotedblright like identity card, or  \textquotedblleft what you know\textquotedblright e.g., password or the personal identification number \cite{R15,R16}. Such traditional biometrics-based systems are successful in many applications like forensics, employee attendance, and mobile or laptop security. Biometrics-based retrieval systems have limited usage in surveillance applications due to the following:

\begin{itemize}
	\item Biometric samples are difficult to capture from video footage of unconstrained environments.
	\item It is challenging to capture physiological samples (e.g., fingerprint) from an uncooperative person.
	\item Biometric attributes (e.g., face) are challenging to acquire due to camera distance and position.
	\item Captured biometrics have poor quality due to low-resolution cameras.
\end{itemize}

\begin{figure*}
	\centering
	\includegraphics[width=0.9\textwidth]{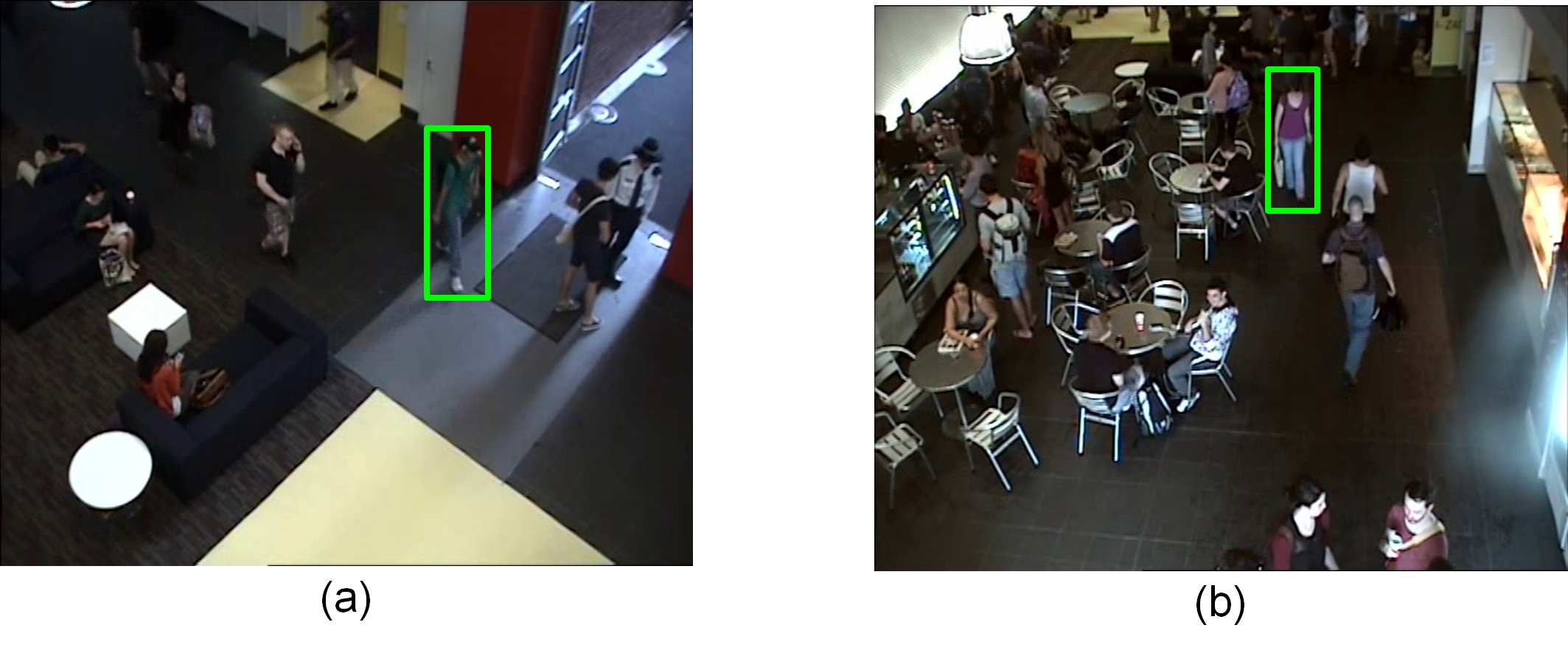}
	\caption{Video surveillance frame samples \cite{R5}.}
	\label{fig:2}       
\end{figure*}

Traditional biometric-based systems fail to retrieve a person from surveillance videos due to the above reasons. Fig. ~\ref{fig:2} shows sample video surveillance frames from the AVSS 2018 challenge II database \cite{R5}. A person of interest is within a green bounding box. It shows various scenarios where traditional biometrics-based systems fail to retrieve the person. Fig. ~\ref{fig:2}(a) shows the environment with poor illumination and a low-resolution frame. Fig. ~\ref{fig:2}(b) shows a scenario with a crowd and a large distance between the camera and the person. Face recognition-based systems fail to retrieve the person in such scenarios. However, attributes like colour and type of clothing, height and gender can help in-person retrieval under such challenging conditions. For example, green (in Fig. ~\ref{fig:2}(a)) and purple (in Fig. ~\ref{fig:2}(b)) coloured clothes are identifiable. Jain et al. \cite{R6} introduce such personal attributes as \textit{soft biometrics}, and many other research articles elaborate on soft biometrics \cite{R6,R7,R8,R9,R10,R11,R12,R13}. Boston Marathon bombing \cite{R14} is a real-world example where police investigation took three days to search out the suspect from hundreds of hours of surveillance videos. Thus, automation of the person retrieval process saves time during a critical investigation. With the ever-increasing demand for security through surveillance, researchers are infusing more interest in developing soft biometric-based person retrieval from low-quality surveillance videos where traditional biometrics fail.

\begin{figure*}
	\centering
	\includegraphics[width=0.9\textwidth]{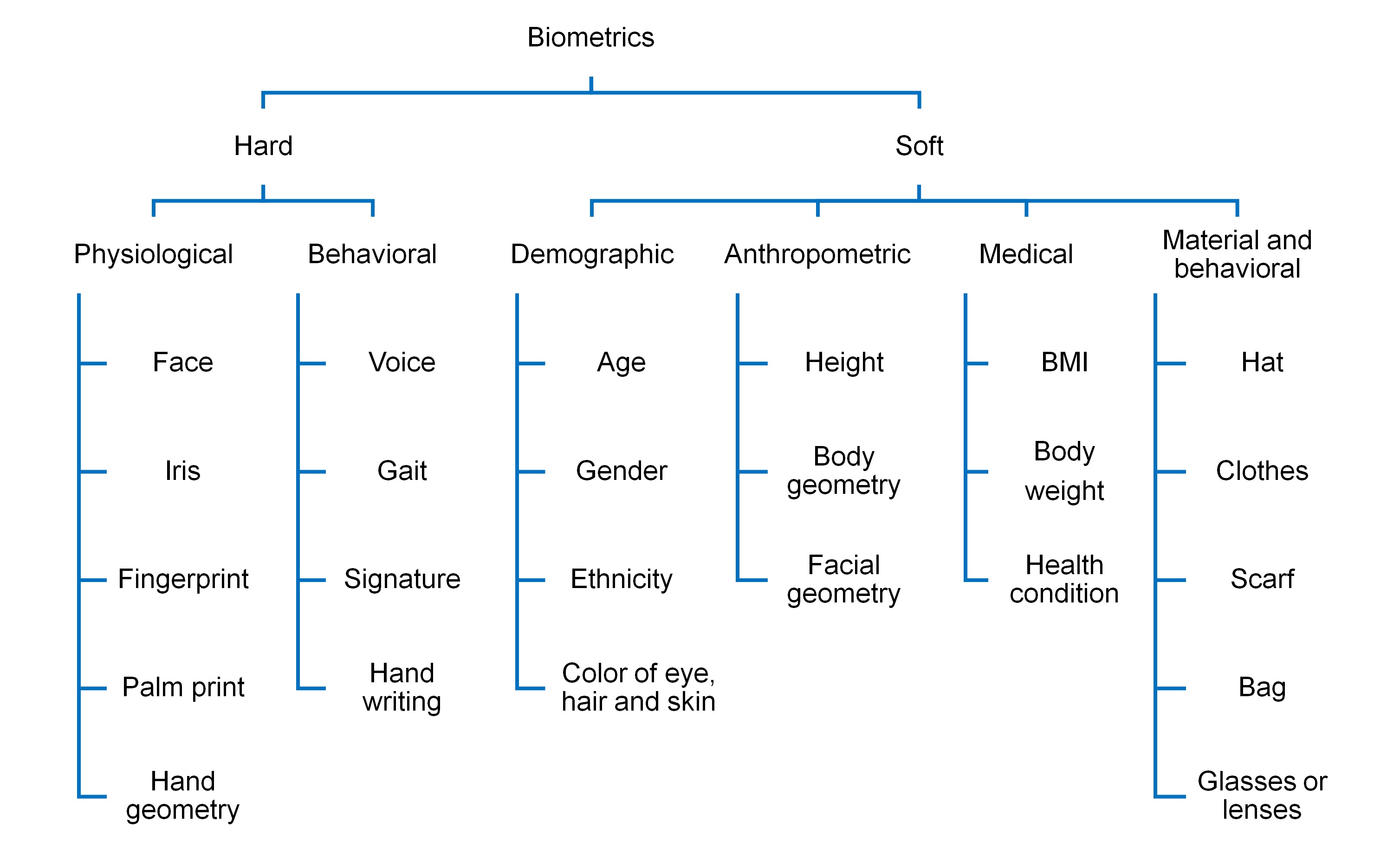}
	\caption{Type of biometrics.}
	\label{fig:3}       
\end{figure*}

Biometrics are broadly classified as \textit{hard} and \textit{soft} biometrics, as shown in Fig. ~\ref{fig:3}. Hard biometrics have physiological and behavioural attributes of a person. Physiologically based methods establish a person's identity using \textit{physiological} characteristics like face, iris, fingerprint, palm print, hand geometry and DNA. \textit{Behavioural} techniques perform authentication by recognizing patterns such as voice, gait, and signature. Although biometrics like face, fingerprint, and palm print are unique to an individual, a single biometric cannot meet the requirements of all applications \cite{R15,R16}. For example, camera-based security systems cannot use fingerprints as a biometric for user authentication. Thus, biometric attribute selection depends on the application. One such application is person retrieval from a distance through a surveillance frame. Here traditional biometrics fail, but soft biometrics play a critical role in retrieval. Soft biometrics are the ancillary information deduced from the human body. Dantcheva et al. \cite{R12} categorized soft biometrics into demographic, anthropometric, medical, material, and behavioural attributes.

The term \textit{demographic} is related to the accumulation and statistical analysis of broad characteristics about groups of people and populations. Age, ethnicity, gender, and race are such characteristics that are used as soft biometrics and also termed as global traits \cite{R17}. Gender, ethnicity and race usually do not show changes over the person's lifespan and hence are useful for search space reduction from surveillance. Gender and age estimation are the most researched demographic attributes. Researchers estimate gender from face \cite{R18,R19,R20,R21,R22,R23,R24,R25,R26}, fingerprint \cite{R27,R28,R29,R30,R31,R32}, iris \cite{R33,R34}, body \cite{R35,R36,R37,R38,R39,R40}, hand \cite{R41,R42,R43,R44,R45,R46} and speech \cite{R47,R48,R49,R50,R51}. Gender from face covers a major research segment and achieves ~99\% \cite{R22} classification accuracy. However, performance decreases dramatically for a realistic and unconstrained environment. It imposes illumination changes, occlusion, different views, and poses. Surveillance videos cover such unconstrained scenarios and it is also very difficult to acquire face, fingerprint, iris, hand, and speech biometrics from surveillance. Gender classification from the full human body is a more suitable way for surveillance applications.

\textit{Anthropometric} measurements are a traditional way of body measurements. Since the era of Bertillonage \cite{R1,R2} they have been in security practices and are currently known as anthropometric and geometric soft biometrics. Facial landmarks related to chin, mouth, eyes, and nose are anthropometric measurements related to facial geometry \cite{R52,R53,R54,R55}. Body height \cite{R56,R57,R58,R59} is the most researched anthropometric attribute. Similarly, other body measurements, like torso length, leg length, and step length \cite{R58}, are also useful.  \textit{Medical} characteristics help monitor a person \cite{R12} with the help of a computer vision-based technique. Bodyweight, body mass index (BMI), and skin quality are soft biometrics used for an application like early detection and prevention of diseases like skin cancer. Another class of soft biometrics is \textit{material and behavioural}. It includes various accessories worn by a person like a hat, scarf, eyeglasses; different clothes and clothing colours; scars, marks, and tattoos on the body. Such soft biometrics also play a crucial role in retrieval. For example, a description, \textit{a male with a green t-shirt and blue jeans wearing a black cap and carrying a backpack}, contains clothing colours, clothing types, accessories (i.e., cap and backpack) as soft biometrics. Material and behavioural attributes are highly time-inconsistent. E.g., a person wearing a white shirt and black pants in the morning may have different clothes (blue t-shirt and white shorts) in the afternoon of the same day. Thus, descriptions for morning and afternoon are different for person retrieval. Time consistency discussion is in Sec.~\ref{sec:2.1}.

\subsection{Characteristics of soft biometrics}
\label{sec:1.2}

Research interest is now more inclined towards applications where person retrieval at a distance is necessary, based on ancillary information like soft biometrics \cite{R60,R61,R62}. Some early research \cite{R6,R63,R64,R65,R66,R67} shows the use of soft biometrics to improve the performance of the primary biometric system by narrowing down the search space in the initial stages. Jain et al. \cite{R6} mentioned that soft biometrics are inexpensive to compute, with no person cooperation requirement, and derived at a distance. Similarly, various pros and cons of soft biometrics are in further discussion.

\paragraph{Registration free / enrollment free:} ~\\
Traditional biometric systems require prior registration of the biometric signature for identification in the future. However, prior registration of soft biometric attributes of a specific person is not required. The training is offline, and the model is useful for retrieval of the individual.

\paragraph{Compatible with human language:} ~\\
Traditional biometrics features like fingerprints are discriminative. However, they cannot be describable by linguistic labels. It creates a semantic gap between human beings and the person retrieval system. Soft biometrics is a form of biometrics that uses labels people generally use to describe or identify each other like \textit{tall, female, young} and \textit{blond hair}. They help to bridge the semantic gap by generating descriptions understandable by humans.

\paragraph{Cost and computationally effective:} ~\\
Face, fingerprint, palm print, and iris-like biometrics require dedicated sensors for acquisition and a constrained environment. Soft biometric attributes like height, gender, and clothing colours are extractable from single low-quality surveillance videos where the traditional biometric acquisition fails. It reduces costs on the overall system development.

\paragraph{Recognizable from a distance:} ~\\
Soft biometrics attributes are identifiable from a distance, e.g., clothing colour, clothing type, gender.

\paragraph{No cooperation from the subject:} ~\\
Soft biometrics attributes (e.g., clothing colour, gender) acquisition does not require any cooperation from the subject. Unconstrained and non-intrusive acquisition is also non-invasive, which makes it suitable for surveillance applications.

\paragraph{Privacy preservation:} ~\\
Soft biometrics-based description (e.g., \textit{a tall female with a blue t-shirt}) only provides partial information. Such descriptions are not unique to an individual. It offers solutions to privacy issues related to soft biometric data storing.

\paragraph{Uniqueness:} ~\\
Soft biometric attributes are not unique to an individual, i.e., a single attribute (e.g., female) does not uniquely retrieve the person. Multiple soft biometrics together can help to overcome this limitation.

\paragraph{Permanence:} ~\\
Soft biometrics are not permanent attributes of an individual like a fingerprint. It makes person retrieval more challenging.

\subsection{Structure of the paper}
\label{sec:1.3}

The paper presents the use of soft biometrics for person retrieval and suggests their strengths and weaknesses. It recommends the most discriminative and robust soft biometrics during person retrieval under challenging conditions. Besides covering the general frame work for retrieval using textual query it also reviews and compares large-scale datasets. It further provides classification for retrieval methods and provides state-of-the-art reviews for soft biometric attribute-based retrieval methods. Further, the integration of benchmark datasets and methods allows an effective comparison of their performance. In this way, the presented review is entirely different from other soft biometrics basics reviews \cite{R9,R10,R11,R12}, which focus mainly on soft biometric attribute retrieval from the person’s image. A review of recent datasets, state-of-the-art retrieval methods including vision and natural language processing, and dataset-method integration is not available in the other review articles.

The contributions of the paper are summarised as follows:

\begin{enumerate}
	\item Recommends the most discriminative soft biometrics under challenging conditions and show case their use for person retrieval.
	
	\item Provides the most comprehensive coverage of soft biometrics datasets.
	
	\item Spans the complete spectrum for state-of-the-art methods; image based, discrete attribute based and natural language description-based retrieval.
	
	\item Integrates datasets and methods for quantitative and objective performance evaluation.
	
	\item Discusses open research problems in person retrieval using soft biometrics and their possible solutions.
\end{enumerate}

\section{Soft biometrics based person retrieval in surveillance videos}
\label{sec:2}

Person retrieval has become the most critical and essential task for surveillance and security. Various government sectors and private organizations invest more in computer vision-based research techniques due to ever-increasing video-based data. It is observable that although the frame in Fig. ~\ref{fig:2}(a) is at low resolution and captured at a distance, a detailed description can still be available. For example, one can get a narrative like \textit{a male with a green t-shirt and blue jeans wearing a black cap, and carrying a backpack}. Such human description naturally contains soft biometrics like gender (\textit{male}), clothing color (\textit{green, blue}), clothing type (\textit{short sleeve}), accessories (\textit{cap, backpack}). It indicates that soft biometrics have a strong relationship with human descriptions. Thus soft biometric attribute-based person retrieval systems are more suitable for an unconstrained environment.

Computer vision has created enormous opportunities to develop an algorithm for intelligent video processing which is useful for retrieval. Developing such algorithms is also tricky due to various problems like video frame resolution, different camera views, poses, illumination conditions, occlusion, and background merging with the foreground.

\subsection{Challenges for person retrieval in surveillance}
\label{sec:2.1}

\paragraph{C1. Low-resolution videos:} ~\\

Surveillance videos have lower resolution since they have to be recording for long periods of time. Fig.~\ref{fig:2} shows such a surveillance frame where it is challenging to extract face information from the low-resolution frame with the distance between the camera and the person. A face recognition-based retrieval system fails to establish identity in such challenging conditions. Also, the low-resolution structure does not help to extract useful features. Hence, it creates a problem for person retrieval.

\paragraph{C2. Illumination changes / varying illumination conditions:} ~\\

\begin{figure*}
	\centering
	\includegraphics[width=0.75\textwidth]{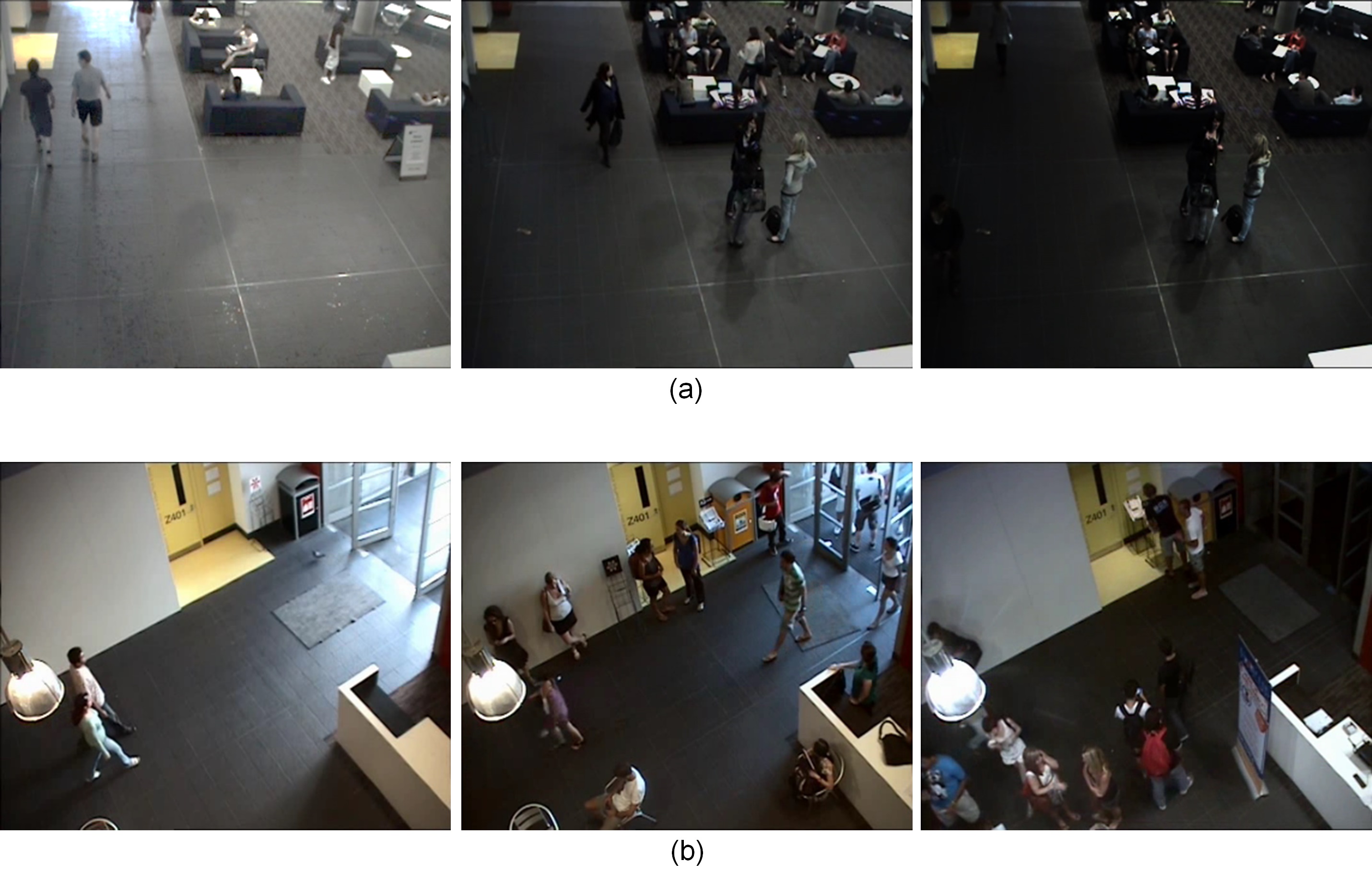}
	\caption{Varying illumination conditions in multiple frames of the single camera. (a) Cam1 and (b) Cam4 in AVSS 2018 challenge II database \cite{R5}. Frames from left to right are with good to poor illumination.}
	\label{fig:4}       
\end{figure*}

The video looks good if the illumination is excellent. A proper lighting condition is the key to better visual information. Fig.~\ref{fig:4}(a) and (b) show single-camera surveillance frames with different illumination conditions. Surveillance frames from left to right are with good to poor illumination conditions. People and their clothing colours are visible in the frames with proper lighting conditions. They are challenging to observe in reduced illumination and create a problem in person detection and colour classification. 

\paragraph{C3. Occlusion:} ~\\

\begin{figure*}
	\centering
	\includegraphics[width=\textwidth]{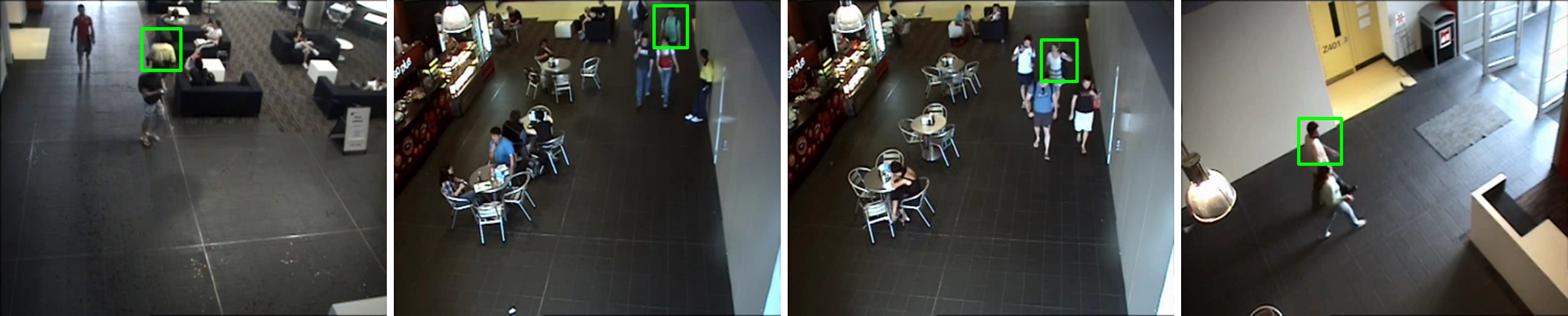}
	\caption{Occlusion in surveillance frames \cite{R5}. The persons of interest are shown with green bounding box.}
	\label{fig:5}       
\end{figure*}

Occlusion in the surveillance frame is a challenge where a person of interest is not completely visible. Fig.~\ref{fig:5} shows such sample frames where a person of interest is within a green bounding box in each frame. It also creates issues for detection and the segmentation process in person retrieval. Occlusion also affects the estimation of an anthropometric attribute like height. 

\paragraph{C4. Merging with background information:} ~\\

\begin{figure*}
	\centering
	\includegraphics[width=\textwidth]{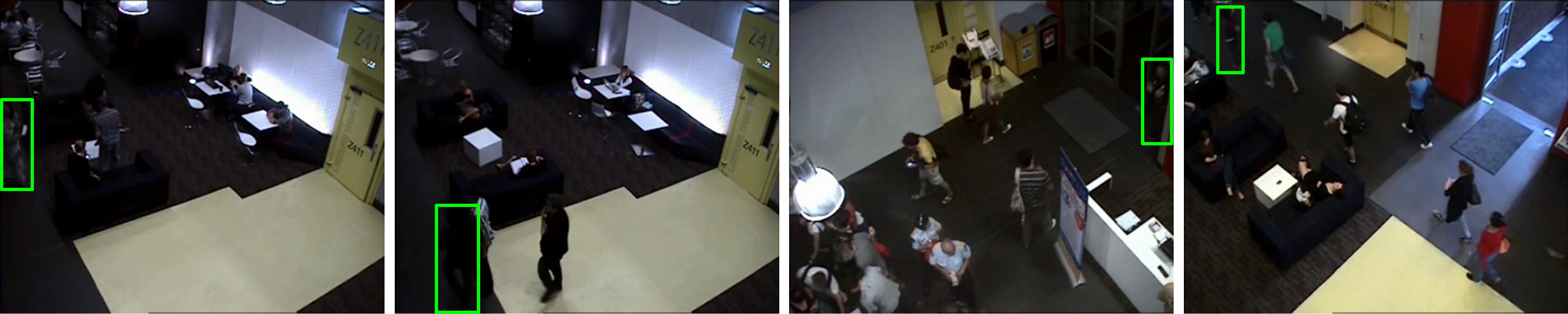}
	\caption{Person of interest merges with background information in surveillance frames \cite{R5}. The persons of interest are shown with green bounding box.}
	\label{fig:6}       
\end{figure*}

Fig.~\ref{fig:6} shows surveillance frames in which the person of interest merges with the background. It happens due to low illumination or similarity in clothing colour and background. Persons are not visible though they are not under occlusion in each frame.

\paragraph{C5. Different illumination in the scene for a single sequence:} ~\\

\begin{figure*}
	\centering
	\includegraphics[width=\textwidth]{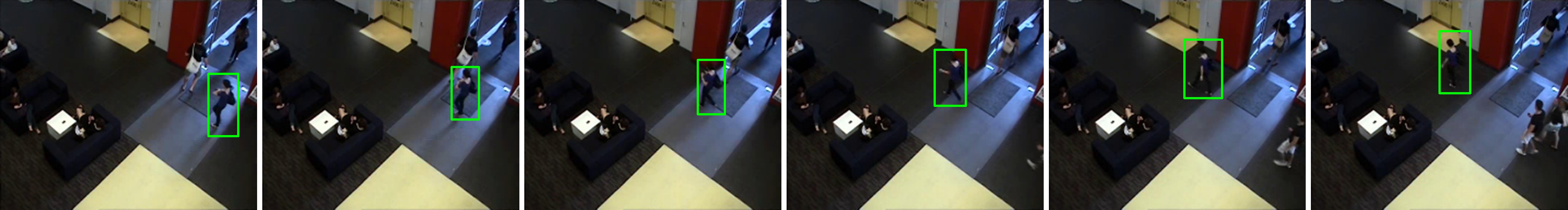}
	\caption{Different lighting condition in a scene for a single sequence \cite{R5}. The person of interest is shown with green bounding box.}
	\label{fig:7}       
\end{figure*}

Fig.~\ref{fig:7} shows a sequence of surveillance frames where illumination varies in different parts of the frame. The person shown within a green bounding box is moving from right to left in the frame. It shows that the outside light is entering the room in the top right corner of the scene. It is observable that the colour information keeps on changing as the person moves through such a region.

\paragraph{C6. Different viewpoints of the same person:} ~\\

\begin{figure*}
	\centering
	\includegraphics[width=0.75\textwidth]{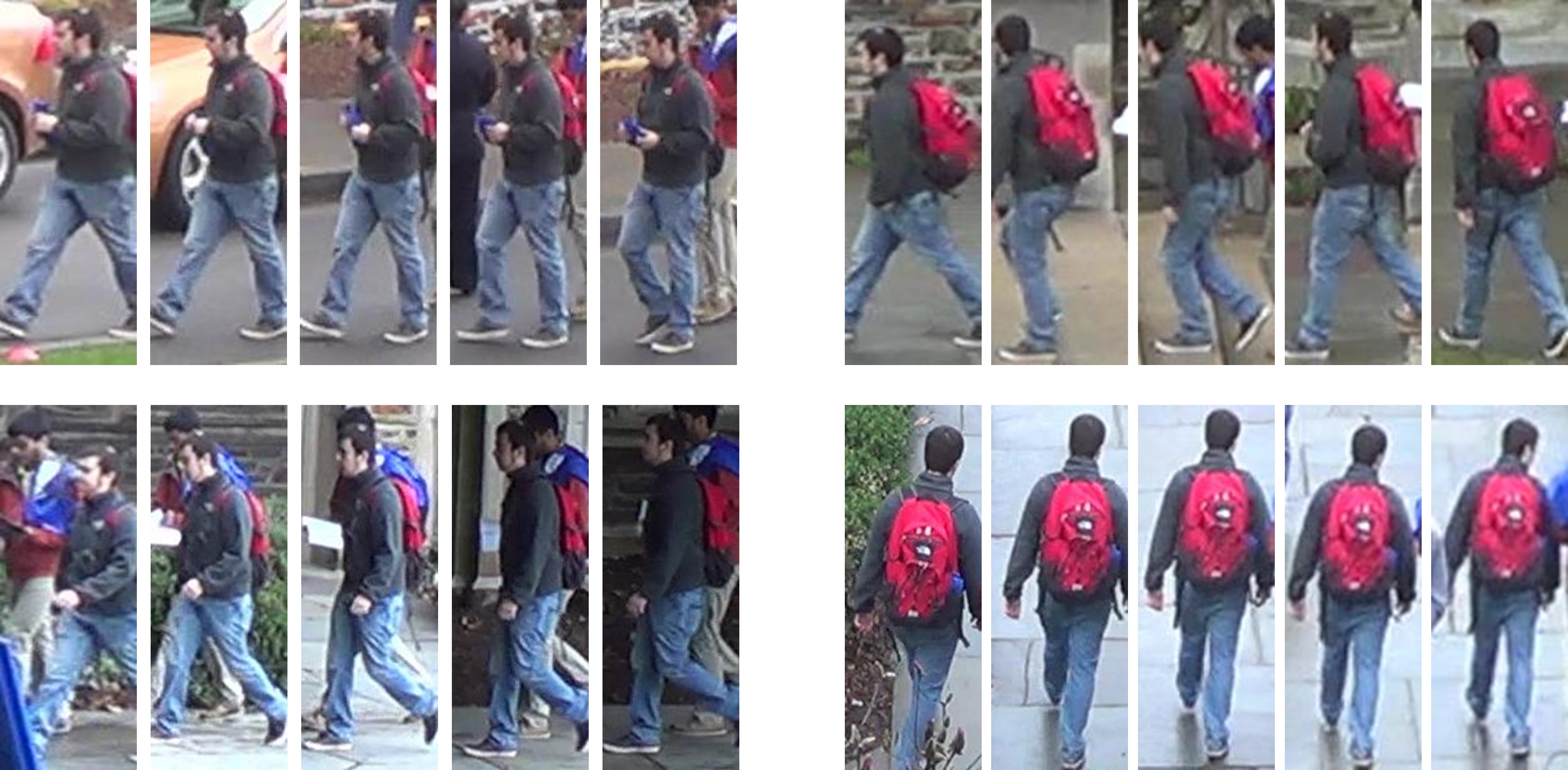}
	\caption{Different viewpoints of same person \cite{R107}.}
	\label{fig:8}       
\end{figure*}

A person may appear with different poses in the same camera scene or scenes of multiple camera networks. Appearance-based methodologies expect minimal or no change in the visual appearance of the person. Fig.~\ref{fig:8} shows a scenario where the pose of a person keeps on changing, and hence the appearance in the camera changes. Retrieval is difficult in such a system. Pose changes also create difficulties for methods that retrieve a person based on his or her gait.

\paragraph{C7. Crowded scene:} ~\\

\begin{figure*}
	\centering
	\includegraphics[width=\textwidth]{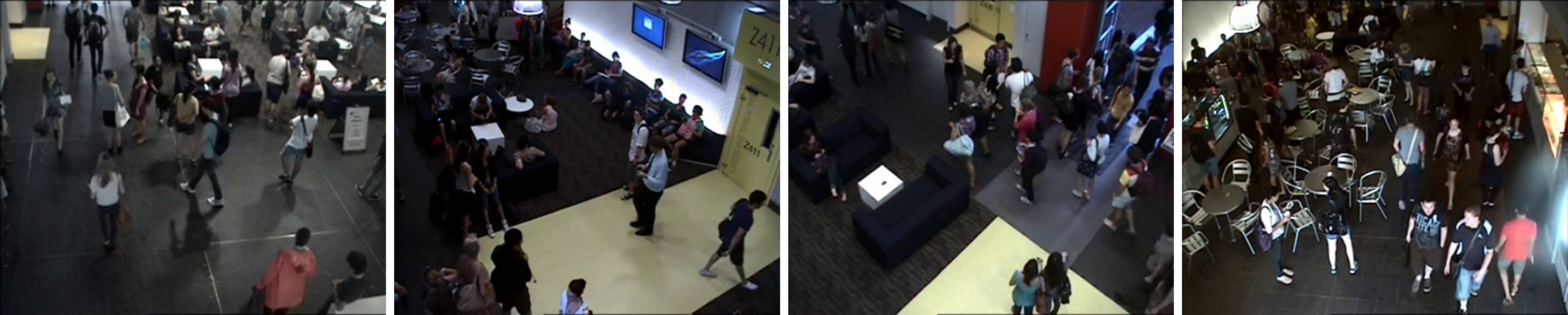}
	\caption{Crowded scene \cite{R5}.}
	\label{fig:9}       
\end{figure*}

Sitting areas, canteens, shopping malls, airports and railway stations are places where a massive crowd is observable. Such frames are in Fig. ~\ref{fig:9}. A crowded scene contains occlusion and persons with a similar appearance cause difficulties in retrieval.

\paragraph{C8. Indoor and outdoor environment:} ~\\

Surveillance cameras are at different indoor places (e.g., shopping malls, canteens) and outdoor places (e.g., university campus, public roads). Fig.~\ref{fig:4}, ~\ref{fig:5}, ~\ref{fig:6}, ~\ref{fig:7} and ~\ref{fig:9} show indoor environments, while Fig.~\ref{fig:8} shows an outdoor environment. Both environments create different illumination conditions. Outdoor environments are affected explicitly by various weather conditions and day-night illuminations.

\paragraph{C9. Time consistency / time duration:} ~\\

As discussed in Sect.~\ref{sec:1.1}, material and behavioural attributes like clothing colour and clothing type are highly time-inconsistent. The same person may have different clothing even on the same day, i.e., morning (white shirt and black pants) and afternoon (blue t-shirt and white shorts). Also, attributes like hairstyle and hair colour may change over the days, and age obviously varies over years.

\paragraph{C10. Type of camera / multiple cameras:} ~\\

Surveillance feeds captured from different types of cameras have other characteristics. For example, red clothing colour perception may differ from a bullet camera to an unmanned aerial vehicle (UAVs) camera. It creates many concerns for clothing colour-based retrieval.

\paragraph{C11. Unavailability of a generic dataset:} ~\\

The available datasets (Sec.~\ref{sec:2.4}) do not cover all the challenges, e.g., a dataset may have only indoor environment images. Also, some of the soft biometric (e.g., clothing colour) annotation may be missing. Such limitations incur the challenge of developing a universal end-to-end system.
\\

The conditions under which a surveillance network operates is challenging. These conditions are responsible for producing unsatisfactory results. The surveillance system should be robust to create a meaningful impact in the real world. It is crucial to select soft biometric attributes that help person retrieval under challenging conditions.

\subsection{Soft biometric attribute selection}
\label{sec:2.2}

Soft biometric attributes are not unique; for example, there may be multiple people with a blue torso colour. Thus, it produces numerous matches for the given textual query. Therefore, it is advantageous to use the most discriminative attributes for person retrieval. Different soft biometrics have certain advantages, and they are discussed below. For example, a surveillance video may contain different view angles and distance. The person's height is invariant to such concerns \cite{R68,R69,R78}. Clothing colour is also one of the most discriminative attributes. It has the following advantages:
\begin{itemize}
	\item Colour is more immune to noise.
	\item It is insensitive to dimension, view angle, and direction.
	\item Colour is recognizable from a far distance.
\end{itemize}

Clothing type is also insensitive to view angle. Gender is identifiable from near as well as far distance and different view angles. Table~\ref{tab:1} shows the strength of soft biometric attributes against challenging conditions. \textquotedblleft \checkmark \textquotedblright  indicates that a particular attribute is useful for the retrieval process against a specific challenging scenario. For example, a standing person's height estimation requires head and feet points \cite{R68,R69,R78}. The extraction of those points is not affected by different view angles, near or far fields, illumination, and low resolution. However, those points are non-extractable when a person is occluded or has a pose like sitting. Similarly, \textquotedblleft $\times$\textquotedblright indicates that a particular attribute is not useful for the retrieval process against the challenge.

\begin{table*}
	\begin{center}
		\caption{Soft biometric attribute selection for person retrieval against different challenges.}
		\label{tab:1}       
		\resizebox{\textwidth}{!}{
			\begin{tabular}{llllllllll}
				\hline\noalign{\smallskip}
				& \textbf{View angles} & \textbf{Far field} & \textbf{Near field} & \textbf{Illumination} & \textbf{Partial occlusion} & \textbf{Pose} & \textbf{Low resolution} & \textbf{Time} & \textbf{Camera} \\
				\noalign{\smallskip}\hline\noalign{\smallskip}
				\textbf{Height} & \checkmark & \checkmark & \checkmark & \checkmark & $\times$ & $\times$ & \checkmark & \checkmark & \checkmark \\
				
				\textbf{Cloth color} & \checkmark & \checkmark & \checkmark & $\times$ & \checkmark & \checkmark & \checkmark & $\times$ & $\times$\\
				
				\textbf{Cloth type} & \checkmark & \checkmark & \checkmark & \checkmark & $\times$ & $\times$ & \checkmark & $\times$ & $\times$\\
				
				\textbf{Gender} & \checkmark & \checkmark & \checkmark & \checkmark & \checkmark & \checkmark & \checkmark & \checkmark & \checkmark\\
				
				\textbf{Torso pattern} & \checkmark & $\times$ & \checkmark & $\times$ & $\times$ & $\times$ & $\times$ & $\times$ & $\times$\\
				
				\textbf{Body geometry} & $\times$ & \checkmark & \checkmark & \checkmark & $\times$ & $\times$ & \checkmark & \checkmark & \checkmark\\
				
				\textbf{Hair (type \& color)} & \checkmark & $\times$ & \checkmark & $\times$ & $\times$ & $\times$ & $\times$ & $\times$ & $\times$\\
				
				\textbf{Shoe (type \& color)} & \checkmark & $\times$ & \checkmark & $\times$ & $\times$ & $\times$ & $\times$ & $\times$ & $\times$\\
				
				\textbf{Accessories (bag, hat)} & $\times$ & $\times$ & \checkmark & $\times$ & $\times$ & $\times$ & $\times$ & $\times$ & $\times$\\
				\noalign{\smallskip}\hline
			\end{tabular}}
		\end{center}
	\end{table*}
	
\subsection{Vision based person retrieval system}
\label{sec:2.3}

This section discusses a person retrieval system that uses a natural language-based textual query. Researchers propose methods using handcrafted feature-based retrieval \cite{R61,R62,R63,R64,R65,R66,R67,R70,R71,R72,R73,R74,R75,R76,R77}, deep learning feature-based linear filtering \cite{R68,R69,R78}, parallel classification of attributes \cite{R79,R80,R81} utilization of Natural Language Processing (NLP) algorithms to process textual queries \cite{R82,R83,R84,R85,R86,R87,R88,R89}. Such a plethora of methods consists of person detection, segmentation, soft biometric attributes classification, and person identification as crucial steps in the person retrieval process. These key steps are given in Fig.~\ref{fig:10} and discussed as follows:

\begin{description}
	\item[\textbf{Step-I (Person detection):}] Person retrieval in surveillance is a challenging task because such scenarios are usually in the wild, containing various objects (e.g., chair, table, car, and train) apart from the person. Hence, person detection is the critical initial step before textual query-based person retrieval. Person detection is the task of locating all persons in the surveillance frame. It was a foreground-background separation problem before the era of deep learning. A person is a foreground object, which is different from the rest of the background. Early research detected the person by different methodologies like background subtraction \cite{R70,R90}, adaptive background segmentation \cite{R91}, Gaussian Mixture Model (GMM) \cite{R92}, and Histograms of Oriented Gradients (HoG) \cite{R93}.
	
	Deep learning-based methodologies are becoming popular in recent years due to their robust feature extraction and learning ability. The computer vision community considers person detection as an object detection problem. Some of the popular object detection frameworks are \textquotedblleft You Only Look Once (YOLO)\textquotedblright \cite{R94}, Single-Shot Multibox Detector (SSD) \cite{R126}, Region-based Convolutional Neural Network (R-CNN) \cite{R95}, Fast R-CNN \cite{R96}, Faster R-CNN \cite{R97} and Mask R-CNN \cite{R98}. Person detection provides a bounding box for each person in the frame.
	\\
	\begin{figure*}
		\centering
		\includegraphics[width=\textwidth]{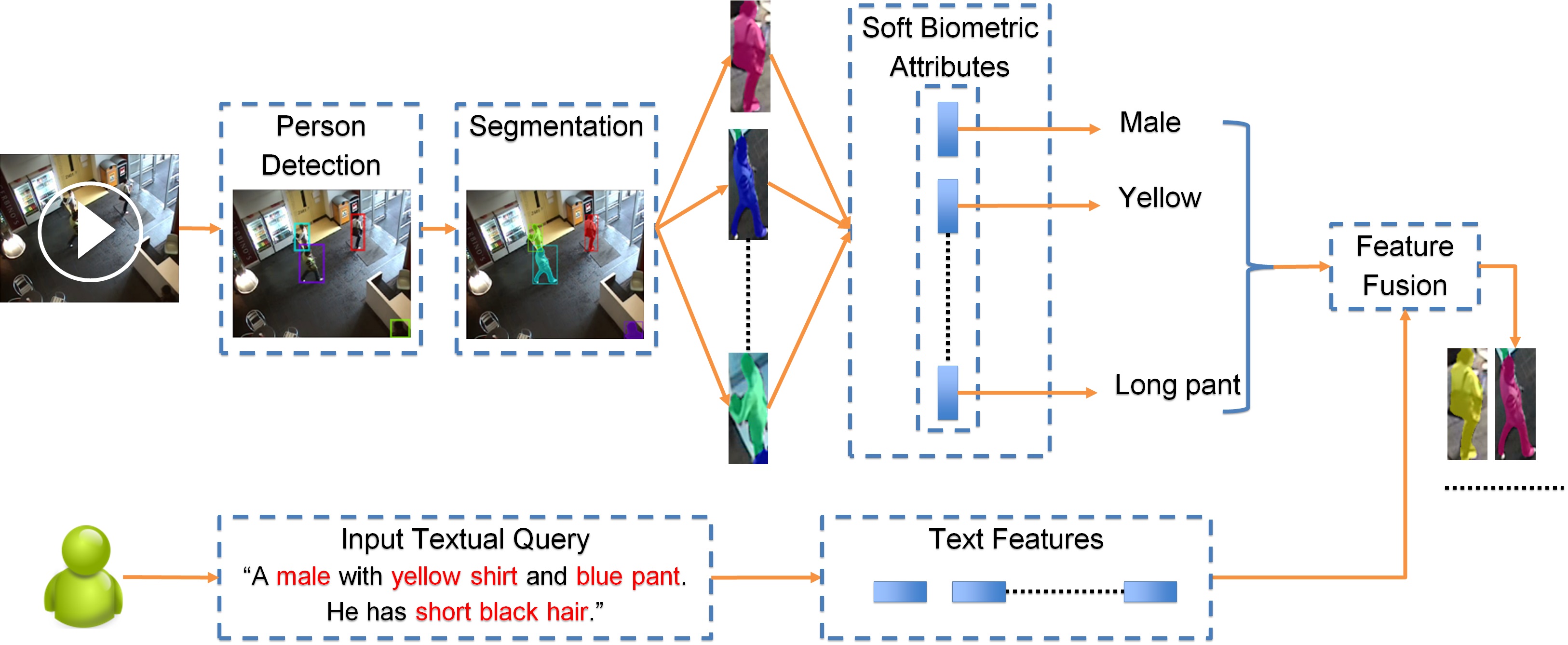}
		\caption{Person retrieval in surveillance using textual query.}
		\label{fig:10}       
	\end{figure*}
	
	\item[\textbf{Step-II (Segmentation):}] Segmentation follows person detection. Segmentation can be either in the form of a body part from a bounding box or semantic segmentation within the bounding box. Such segmentation is shown in Fig.~\ref{fig:11}. A full-body can be segmented into three major parts: head and shoulders, upper body, and lower body (see Fig.~\ref{fig:11}(a)). Fig.~\ref{fig:11}(b) shows semantic segmentation in which each pixel belonging to the person within the bounding box has a label. Mask R-CNN \cite{R98} provides both person detection and semantic segmentation together. \\
	
	\begin{figure}
		\centering
		\includegraphics[width=\columnwidth]{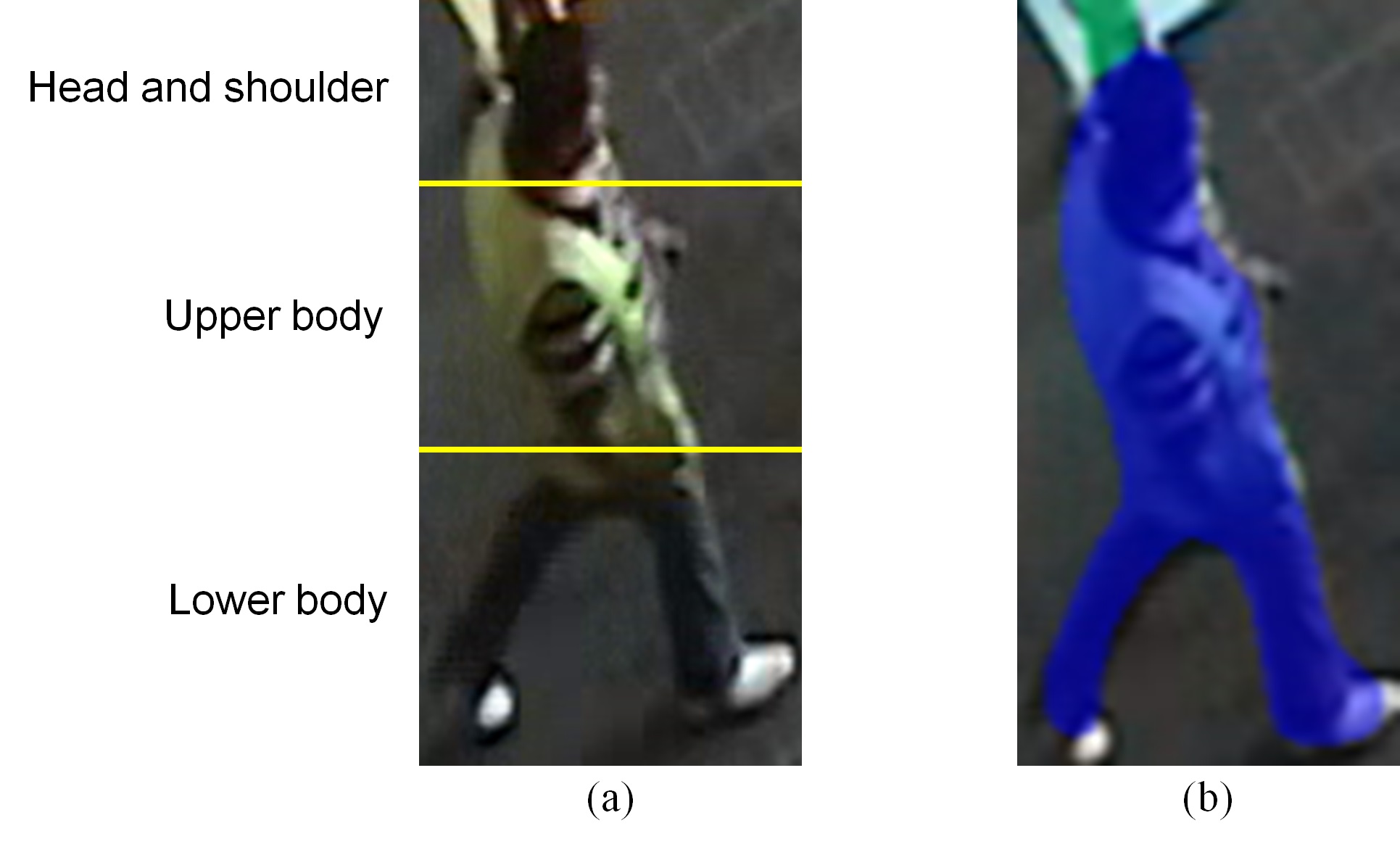}
		\caption{Segmentation: (a) body part segmentation and (b) semantic segmentation.}
		\label{fig:11}       
	\end{figure}
	
	\item[\textbf{Step-III (Soft biometric attribute recognition):}] Full-body images and segmentation outputs are supplied to extract soft biometric attributes. Full-body images are useful for extracting characteristics like height and gender \cite{R68,R69,R78}. On the other hand, clothing colour, clothing texture, clothing type, shoes, hair, and accessory attributes are available from different body parts. Usually, details from video frames are visual features. Recent development shows multi-attribute learning for person attribute recognition. Researchers propose various networks like Deep learning-based Multiple Attribute Recognition (DeepMAR) \cite{R152}, Attribute Convolutional Net (ACN) \cite{R153}, and Multi-Label Convolutional Neural Network (MLCNN) \cite{R154}. \\
	
	\begin{figure}
		\centering
		\includegraphics[width=\columnwidth]{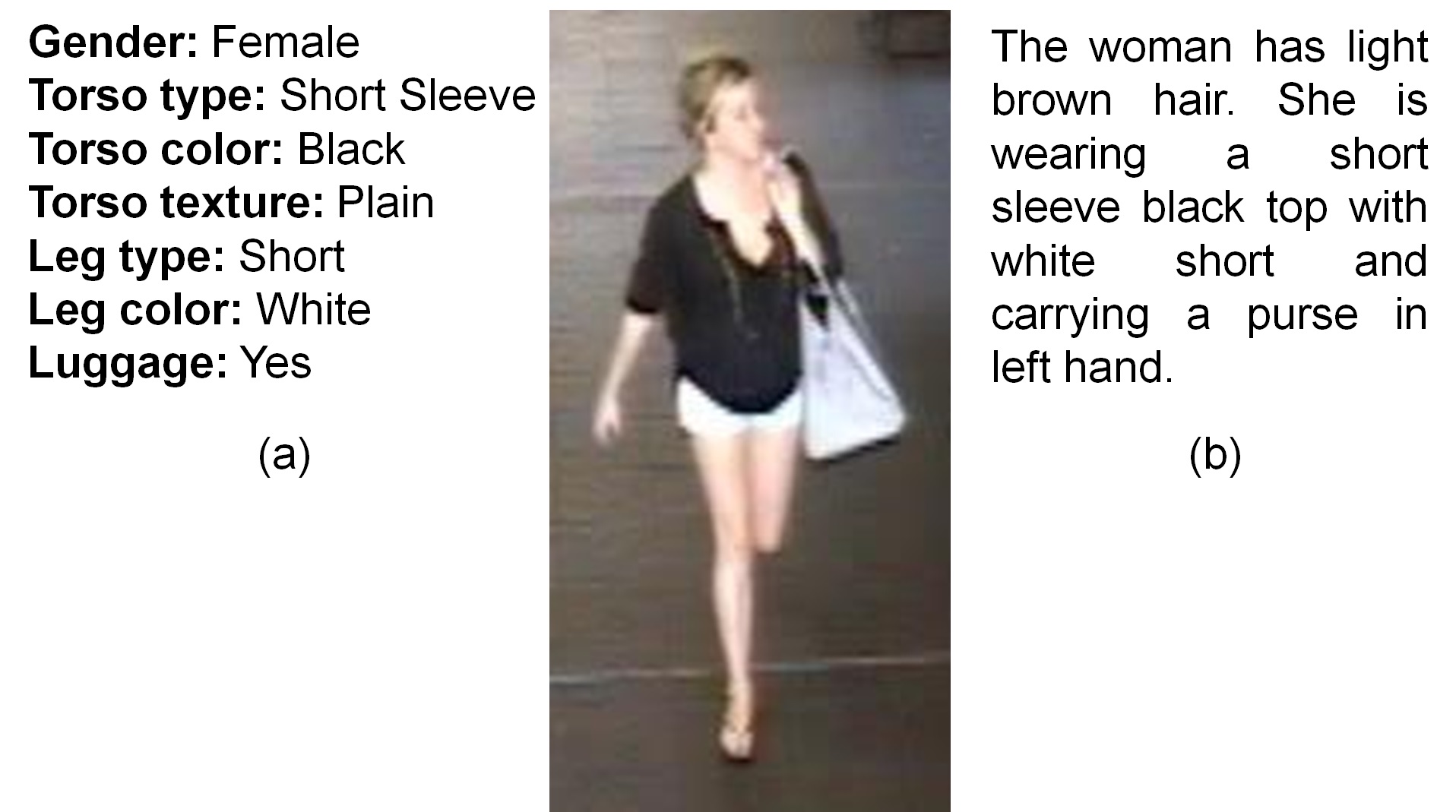}
		\caption{Textual query: (a) discrete attribute query and (b) natural language description}
		\label{fig:12}       
	\end{figure}
	
	\item[\textbf{Step-IV (Text feature extraction):}] Textual attribute query or natural language description is another input to the person retrieval system. Such textual query samples are given in Fig.~\ref{fig:12}. Textual attribute query (Fig.~\ref{fig:12}(a)) is cheaper to collect in terms of attribute wise information and has a less complicated sentence structure in comparison with natural language description (Fig.~\ref{fig:12}(b)). The data is collected separately for all attributes which is useful in the retrieval process. For example, torso type attribute value is available from one of the predefined classes  \{Long Sleeve, Short Sleeve, No Sleeve\}. However, such a discrete attribute query has a fragile appearance, descriptive ability, and practical usage limitation.
	
	In contrast to textual attribute query, natural language description has more complex sentence structures. However, it provides a detailed description of the person. It is essential to extract relevant information from such a human verbal description. For example, the description in Fig. ~\ref{fig:12}(b), \textquoteleft women\textquoteright, \textquoteleft brown hair\textquoteright, \textquoteleft blacktop\textquoteright are more relevant information than \textquoteleft the\textquoteright, \textquoteleft a\textquoteright, and \textquoteleft has\textquoteright. Such relevant information forms textual features. NLP based algorithms help to extract textual features from natural language descriptions. \\
	
	\item[\textbf{Step-V (Feature fusion and retrieval):}] Visual features for each person are available from surveillance videos, and textual features are extractable from the textual query. Person retrieval from surveillance using textual query covers two major problem domains: (i) computer vision and (ii) natural language processing. Hence, cross-modal feature embedding is applicable in the feature fusion block (see Fig.\ref{fig:10}). Finally, a person(s) matching the textual query is retrieved. There is a possibility of multiple person retrieval as soft biometrics are not unique to an individual.
	
\end{description}

The accuracy of person retrieval depends on the complexity of the training and testing data sets. The retrieval robustness relies on the availability of the richly annotated data sets. The performance evaluation of the method is done by evaluation metrics like accuracy, True Positive Rate (TPR) and Intersection-over-Union (IoU). Therefore, the following sub-sections discuss person retrieval datasets (available in the public domain) and evaluation metrics.

\subsection{Person retrieval datasets}
\label{sec:2.4}

A variety of challenging datasets are an essential entity for any research and development task. Datasets with extensive annotations and reliable evaluation strategies help to create a robust algorithm. Researchers have developed many public datasets for person identification and re-identification in the past two decades. Table~\ref{tab:2} shows only datasets with soft biometric annotations, which help create textual attribute query or natural language description for person retrieval. 

Dataset comparisons have the following parameters: number of persons, number of images, resolution of the image (width $\times$ height)), number of soft biometrics annotations, number of cameras used to collect the dataset, type of attribute query, challenges covered, and whether the dataset is with full surveillance frame or cropped person. Abbreviations for challenges and types of attributes are in Table~\ref{tab:2}, which are useful for further discussion. Layne et al. \cite{R102} annotate the VIPeR \cite{R99} dataset with 15 soft biometric attributes like \textit{shorts, skirt, sandals, backpack, jeans, logo, v-neck, stripes, sunglasses, open-outerwear, headphones, long-hair, short-hair, gender} and \textit{carrying an object}. These are binary attributes e.g., gender has value from set \{0, 1\}, where 0 = male and 1 = female.

Zhu et al. \cite{R110} introduce the Attributed Pedestrians in Surveillance (APS) database specifically for pedestrian attribute classification, which can also be useful for person retrieval. The APS consists of 11 binary attributes and two multi-class (total 35) attributes, i.e., clothing colour of upper-body and lower-body. APS is the first dataset to use clothing colour and covers more challenges compared to VIPeR. Layne et al. \cite{R103} further extend annotations for VIPeR \cite{R99}, PRID \cite{R100}, and GRID \cite{R101} from 15 to 21 by introducing shirt colour (red, blue, green), shirt shade (light and dark) and pant and hair colour. New datasets like VIPeR and PRID are having fixed image resolution, limited cameras, \textit{C2, C5}, and \textit{C8}. GRID upgrades such parameters and introduces \textit{C1} and \textit{C3}. However, it is limited to some images. VIPeR, PRID, and GRID contain photos from outdoor scenes only.

\begin{landscape}
	\begin{table}
		\caption{Datasets with soft biometrics annotations.}
		\label{tab:2}
		\begin{tabular}{p{0.12\columnwidth}p{0.06\columnwidth}p{0.08\columnwidth}p{0.08\columnwidth}p{0.08\columnwidth}p{0.08\columnwidth}p{0.08\columnwidth}p{0.06\columnwidth}p{0.08\columnwidth}p{0.08\columnwidth}}
			\hline\noalign{\smallskip}
			\textbf{Dataset} & \textbf{Year} & \textbf{\# Person} & \textbf{\# Images} & \textbf{Resolution (W $\times$ H)} & \textbf{\# Soft attributes} & \textbf{\# Camera} & \textbf{DA / NLD} & \textbf{Challenges} & \textbf{Full frame / cropped person} \\
			\noalign{\smallskip}\hline\noalign{\smallskip}
			
			VIPeR$^{a}$ \cite{R99,R102} & 2012 & 632 & 1,264 & 48 $\times$ 128 & 15 & 2 & DA & C2, C6, C8, C10 & Cropped \\
			
			APiS \cite{R110} & 2013 & \textendash & 3.661 & 48 $\times$ 128 & 35 & \textendash & DA & C2, C3, C6, C7, C8 & Cropped \\
			
			VIPeR$^{b}$ \cite{R99,R103} & 2014 & 632 & 1,264 & 48 $\times$ 128 & 21 & 2 & DA & C2, C6, C8, C10 & Cropped \\
			
			PRID \cite{R100,R103} & 2014 & 934 & 24,541 & 64 $\times$ 128 & 21 & 2 & DA & C2, C6, C8, C10 & Cropped \\
			
			GRID \cite{R101,R103} & 2014 & 1,025 & 1,275 & from 29 $\times$ 67 to 169 $\times$ 365 & 21 & 8 & DA & C1, C2, C3, C6, C8, C10 & Cropped \\
			
			PETA \cite{R104} & 2014 & \textendash & 19,000 & from 17 $\times$ 39 to 169 $\times$ 365 & 105 & \textendash & DA & C1, C2, C4, C6, C8 & Cropped \\
			
			Mareket-1501 \cite{R105} & 2015 & 1,501 & 32,217 & 64 $\times$ 128 & 30 & 6 & DA & C2, C3, C6, C8, C10 & Cropped \\
			
			SoBiR \cite{R106} & 2016 & 100 & 1,600 & from 60 $\times$ 150 to 191 $\times$ 297 & 12 & 8 & DA & C1, C2, C6, C8, C10 & Cropped \\
			
			DukeMTMC-reID \cite{R107} & 2017 & 1,812 & 36,441 & from 34 $\times$ 85 to 193 $\times$ 477 & 23 & 8 & DA & C1, C2, C3, C6, C8, C10 & Cropped \\
			
			CUHK-PEDES \cite{R85} & 2017 & 13,003 & 40,206 & from 13 $\times$ 34 to 374 $\times$ 800 & \textendash & \textendash & NLD & C1, C2, C3, C6, C8 & Cropped \\
			
			PA-100k \cite{R117} & 2017 & \textendash & 1,00,000 & from 50 $\times$ 100 to 758 $\times$ 454 & 26 & 598 & DA & C1, C2, C6, C8, C10 & Cropped \\
			
			RAP \cite{R108} & 2018 & 2,589 & 84,928 & from 33 $\times$ 81 to 415 $\times$ 583 & 72 & 25 & DA & C1, C2, C3, C4, C6, C7, C8, C10 & Cropped \\
			
			AVSS 2018 Challenge II \cite{R5} & 2018 & 151 & 20,453 (frames) & 704 $\times$ 576 & 16 & 6 & DA & C2, C3, C4, C5, C6, C7, C8, C9, C10 & Full frame \\
			
			RAP-LSPRC \cite{R109} & 2019 & 2,589 & 67,571 & from 33 $\times$ 81 to 415 $\times$ 583 & 72 & 25 & DA & C1, C2, C3, C4, C6, C7, C8, C10 & Cropped \\
			
			P-DESTRE \cite{R155} & 2020 & 269 & \textendash & 3840 $\times$ 2160 & 16 & \textendash & DA & C2, C3, C6, C7, C8, C9 & Full frame \\
			\noalign{\smallskip}\hline\noalign{\smallskip}
			
			\multicolumn{10}{p{0.82\columnwidth}}{\textbf{Abbreviations used:}} \\
			
			\noalign{\smallskip}
			
			\multicolumn{4}{p{0.34\columnwidth}}{$^{a}$ Layne et al. \cite{R102} introduce soft biometric annotation to VIPeR dataset in 2012.} & \multicolumn{3}{p{0.24\columnwidth}}{C1 \textendash Different resolutions} & \multicolumn{3}{p{0.22\columnwidth}}{C6 \textendash Different viewpoint / poses.} \\
			
			\multicolumn{4}{p{0.34\columnwidth}}{$^{b}$ Layne et al. \cite{R103} further added 6 more soft biometric annotation to VIPeR dataset in 2014.} & \multicolumn{3}{p{0.24\columnwidth}}{C2 \textendash Varying illumination conditions} & \multicolumn{3}{p{0.22\columnwidth}}{C7 \textendash Crowded scene.} \\
			
			\multicolumn{4}{p{0.34\columnwidth}}{W = Width of an image, H = height of an image.} & \multicolumn{3}{p{0.24\columnwidth}}{C3 \textendash Occlusion} & \multicolumn{3}{p{0.22\columnwidth}}{C8 \textendash In/outdoor environments} \\
			
			\multicolumn{4}{p{0.34\columnwidth}}{DA = Discrete Attribute textual query.} & \multicolumn{3}{p{0.24\columnwidth}}{C4 \textendash Merging with background} & \multicolumn{3}{p{0.22\columnwidth}}{C9 \textendash Time consistency/duration} \\
			
			\multicolumn{4}{p{0.34\columnwidth}}{NLD = Natural Language Description.} & \multicolumn{3}{p{0.24\columnwidth}}{C5 \textendash Different illumination in the scene for a single sequence} & \multicolumn{3}{p{0.22\columnwidth}}{C10 \textendash Type of camera / multiple camera} \\
			
			\noalign{\smallskip}\hline\noalign{\smallskip}
			
		\end{tabular}
	\end{table}
\end{landscape}

Deng et al. \cite{R104} introduce the first large-scale PEdesTrian Attribute (PETA) dataset to overcome the limitations of VIPeR, APS, PRID, and GRID datasets. They introduce \textit{C1, C2, C4, C5, C7, C8} and diversity by annotating 61 binary and 4 multiclass (total 105) attributes in 19,000 images collected from surveillance datasets like 3DpeS \cite{R111}, CAVIAR4ReID \cite{R112}, CUHK \cite{R113}, GRID \cite{R101}, i-LIDS \cite{R114}, PRID \cite{R100}, SARC3D \cite{R115}, TownCentre \cite{R116} and VIPeR \cite{R99}. Martinho et al. \cite{R106} create the SoBiR dataset with 12 multi-class attributes. It is a comparatively smaller dataset with 1,600 images. However, each soft biometric attribute possesses multiple classes which help discriminate against people better.

Recent research advancement is increasing rapidly in deep learning domains, which requires a more diverse and huge amount of data for better generalization. Datasets like VIPeR, APS GRID, and SoBiR are too small for deep learning-based frameworks. Datasets published in the last five years provide a good amount of data to create robust models and algorithms. Large-scale datasets such as Mareket-1501 \cite{R105} have 32,217 images of 1,501 persons and DukeMTMC-reID \cite{R107} provides 36,441 images of 1,812 persons. These small and large-scale datasets use discrete attribute-based queries (Fig.~\ref{fig:12}(a)), which has limited practical usage, as discussed in Sect.~\ref{sec:2.3}. Thus, Li et al. \cite{R85} propose a large-scale dataset with natural language description-based annotations. The dataset is known as CUHK Person Description Dataset (CUHK-PEDES). It provides 40,206 images of 13,003 persons with 80,412 description sentences. It is the only large-scale dataset with a natural language description that provides an opportunity to explore the relationship between language and vision for a person retrieval problem.

Liu et al. \cite{R117} propose a large-scale pedestrian attribute (PA) dataset with 100000 images. It is known as PA-100k. It consists of the highest number of images in all the datasets with a more diverse collection from 598 scenes. It can only be used for attribute-based person retrieval, not for person re-identification. Further increasing the scale in all aspects, Li et al. \cite{R108} created a Richly Annotated Pedestrian (RAP). Li et al. \cite{R109} also make a RAP - Large-Scale Person Retrieval Challenge (RAP-LSPRC) dataset. RAP-LSPRC is a subset of RAP with a comprehensive evaluation and benchmark dataset. By far, RAP contains the highest number of images (84,928) collected from 25 surveillance cameras installed in the unconstrained indoor environment of a shopping mall. RAP and RAP-LSPRC are useful for both person retrieval and person re-identification tasks. 

Datasets discussed so far contain an image gallery of cropped persons from surveillance video frames. Hence, in such image problems related to person detection, occlusion, merging with background, crowded scene, and illumination variations in a single sequence cannot be overcome. It limits the development of an end-to-end intelligent surveillance system. Halstead et al. first developed an AVSS 2018 challenge II dataset with full surveillance frames (20,453) collected from 6 indoor cameras. It is the only dataset that contains video sequences of 151 persons (110 training + 41 testing) with varying surveillance frames from 21 to 290. They provide shallow resolution (704 $\times$ 576)) frames. They cover most real-time surveillance challenges (Sect.~\ref{sec:2.1}) to develop an end-to-end solution for person retrieval. Such surveillance frames are in Fig.~\ref{fig:4}, ~\ref{fig:5}, ~\ref{fig:6}, ~\ref{fig:7}, and Fig. ~\ref{fig:9}.

Recently, Kumar et al. \cite{R155} released a UAV-based dataset for Pedestrian Detection, Tracking, Re-Identification, and Search (P-DESTRE) from aerial devices. It also contains full surveillance frame videos. These videos are captured with 4K spatial resolution (3840 $\times$ 2160) and over different days and times. Thus, P-DESTRE is the first dataset to provide consistent ID annotations across multiple days. Other datasets discussed so far do not cover this challenge. Although this dataset covers rich annotations, it lacks clothing colour annotations. By reviewing all the datasets, the commonly used soft biometric attribute set is  \{\textit{gender, upper-body cloth type, lower-body cloth type, upper-body cloth color, lower-body cloth color, upper-body cloth texture, clothing style, height, hair color, hairstyle, backpack, carrying an object, eye-wear, shoe}\}. Such variety and diversity show the continuous evolution of challenging datasets proposed by active researchers in the vision and language fields.

\subsection{Evaluation metric}
\label{sec:2.5}

Different performance measures evaluate the performance of any methodology. The section covers such evaluation metrics before discussing different procedures (Sec.~\ref{sec:3}) because knowledge of metrics helps to understand the performance of the method better.

\paragraph{Accuracy (ACC):} ~\\
Classification accuracy usually evaluates machine learning models. It is the ratio of correct predictions to the total number of samples in the test. This metric represents the overall efficiency of the system. The metric works well with a balanced dataset, i.e., the number of positive and negative samples in the dataset are equal. The accuracy equation is given by eq.~\ref{ACC},

\begin{equation}
\label{ACC}
ACC=\frac{TP+TN}{P+N}=\frac{TP+TN}{TP+TN+FP+FN}
\end{equation}

Where,

\begin{description}
	\item[P] = The number of real positive samples in the data
	\item[N] = The number of real negative samples in the data
	\item[TP] = True positive i.e., correctly identified
	\item[FP] = False positive i.e., incorrectly identified
	\item[TN] = True negative i.e., correctly rejected
	\item[FN] = False negative i.e., incorrectly rejected
\end{description}

Let us consider an example of gender classification for 100 persons to understand the metric and the same example is going to be used for all the metric discussions in this section. Here, the positive class is male and negative class is female.


\begin{table}[]
	\begin{tabular}{p{0.45\columnwidth}p{0.45\columnwidth}}
		
		\multicolumn{2}{p{0.9\columnwidth}}{\textbf{Gender classification example:}} \\
		
		\noalign{\smallskip}
		
		\cellcolor[HTML]{FABF8F} \textbf{True positive (TP):}{\smallskip} \newline Class: Male\newline Model predicted: Male\newline Number of TP results: 1 & \cellcolor[HTML]{92D050} \textbf{False Positive (FP):}{\smallskip} \newline Class: Female\newline Model predicted: Male\newline Number of FP results: 1 \\
		
		\cellcolor[HTML]{92D050} \textbf{False Negative (FN):}{\smallskip} \newline Class: Male\newline Model predicted: Female\newline Number of FN results: 10 & \cellcolor[HTML]{FABF8F} \textbf{True Negative (TN):}{\smallskip} \newline Class: Female\newline Model predicted: Female\newline Number of TN results: 88
	\end{tabular}
\end{table}

\begin{equation*}
ACC = \frac{1+88}{1+88+1+10} = 0.89
\end{equation*}

Accuracy is 89\% i.e., 89 correct predictions out of 100 persons. Out of 100 persons, 11 are male, and 89 are female. 88 females out of 89 are correctly identified. But, only one male out of 11 is correctly identified. Thus, 89\% of model accuracy seems very good. But accuracy alone as a metric can't help such a class-imbalanced problem as it gives a false sense of high accuracy. A real problem can arise when the cost of misclassifying minor samples is very high. Thus, the metric discussed below is helpful, along with accuracy.

\paragraph{Precision:} ~\\
It is a positive predictive value and decides the fraction of correct positive predictions in total positive predictions. Thus, it provides the proportion of positive predictions of the samples that are correct. It is a measure of quality, and its high value means that the classifier returns more relevant results, i.e., it suggests how many selected items are appropriate. Precision is calculated as eq.~\ref{precision},

\begin{equation}
\label{precision}
Precision = \frac{TP}{TP+FP}
\end{equation}

For gender classification example,

\begin{equation*}
Precision = \frac{1}{1+1} = 0.5
\end{equation*}

The model has precision of 0.5 i.e., 50\% of the time the model is correct when it predicts the person gender as male.

\paragraph{Recall or True Positive Rate (TPR):} ~\\
It represents the classifier's sensitivity, and it is the fraction of correct positive predictions to the total number of relevant samples. It provides the proportion of positive predictions in the correctly retrieved samples. It is a measure of the quantity and suggests how many pertinent samples have been selected. Recall is calculated as eq.~\ref{TPR},

\begin{equation}
\label{TPR}
Recall \; or \; TPR = \frac{TP}{TP+FN}
\end{equation}

For gender classification example,

\begin{equation*}
Recall = \frac{1}{1+10} = 0.09
\end{equation*}

The model has a recall of 0.09 i.e., it correctly identifies 9\% of total male persons.

%
%
%
%
%
%
%

\paragraph{Mean Average Precision (mAP):} ~\\
It is an average of maximum precisions at different recall values. An object detector accuracy is measured by mAP. It is calculated as eq.~\ref{mAP},

\begin{equation}
\label{mAP}
mAP=\frac{1}{Total \; Samples} \sum_{Recall_{i}} Precision(Recall_{i})
\end{equation}

%

\begin{figure}
	\centering
	\includegraphics[width=\columnwidth]{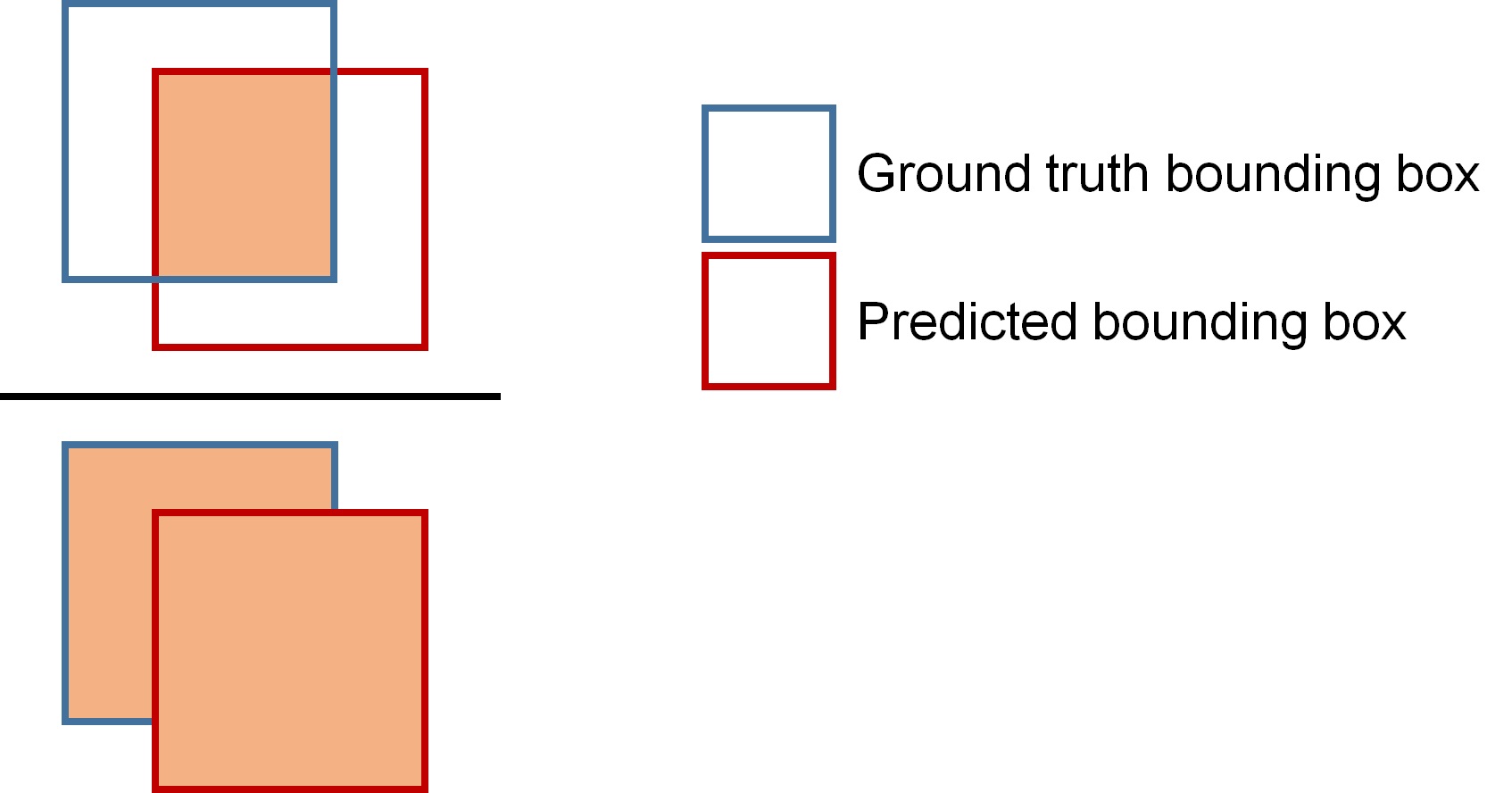}
	\caption{Intersection-over-Union (IoU).}
	\label{fig:13}       
\end{figure}

\paragraph{Intersection-over-Union (IoU):} ~\\
The localisation accuracy of the object detector is measurable by Intersection-over-Union (IoU). It uses the bounding box of the object for evaluation. IoU is the fraction of the intersecting area to the union area between the output bounding box and the ground truth. It measures how good the detector is in localizing objects to the ground truth. Fig.~\ref{fig:13} depicts the IoU metric.

\begin{equation}
\label{IoU}
IoU=\frac{Area \; of \; overlap}{Area \; of \; union}=\dfrac{D\cap GT}{D\cup GT}
\end{equation}

$D$ = bounding box output of algorithm and $GT$ = ground truth bounding box.

Different person retrieval methodologies reviewed below use the evaluation metrics discussed above. Methods in \cite{R69,R74,R75,R76,R78,R79,R80} use IoU or average IoU, \cite{R69,R70,R78} use TPR and \cite{R81,R82,R84,R85,R89,R134,R135,R136,R137,R138,R139} use Top-1 accuracy as evaluation metrics.

\section{Person retrieval methodologies}
\label{sec:3}

Person retrieval is a subject undergoing intense study for the past two decades due to its prime application to public safety. A large-scale intelligent surveillance system requires person retrieval using visual attributes or available person image(s). Vision community researchers are proposing a plethora of methodologies for robust person retrieval. They have two major categories based on the type of input query:

\begin{itemize}
	\item Image-based retrieval.
	\item Soft biometric attribute-based retrieval.
\end{itemize}

\begin{figure*}
	\centering
	\includegraphics[width=\textwidth]{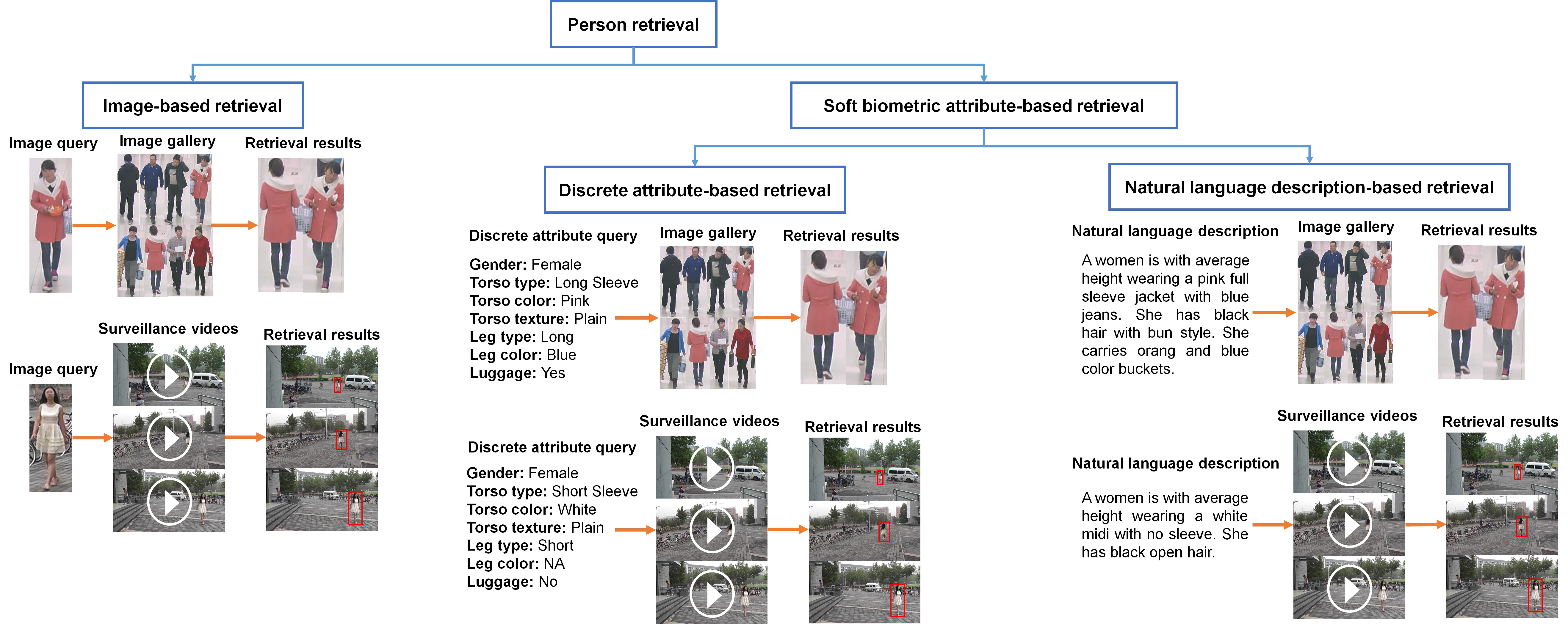}
	\caption{Classification of person retrieval methodologies based on the type of input query. Each classification category shows example of person retrieval from cropped person image gallery and full frames of surveillance video. Images shown in image gallery are adopted from RAP \cite{R108} dataset and surveillance video frames are adopted from PRW \cite{R118} dataset to showcase the example.}
	\label{fig:14}       
\end{figure*}

Classification of person retrieval methodologies based on the type of input query is shown in Fig.~\ref{fig:14}. A person is retrievable from the cropped person image gallery or full frames of surveillance video. The latter is more suitable for real-time scenarios. The person retrieval example of each category is also shown in Fig.~\ref{fig:14}.

Image-based person retrieval methodologies aim to retrieve the target from the large-scale image gallery. This large-scale image gallery or dataset is from different non-overlapping cameras. Such person retrieval techniques are known as \textit{person re-identification} (Re-ID). These techniques require at least one image of the target person for querying the system to perform re-identification of the same identity from different cameras.  A person ReID technique assumes that the image gallery contains at least one image of the target person captured from a camera other than the query image. Such methodologies fail when a probe image is not available.

In contrast, soft biometric attribute-based person retrieval methodologies do not require a probe image. It uses a semantic person description generated from soft biometric attributes. Such a description (textual query) is input to the person retrieval system. Soft-biometric attribute-based retrieval is further divisible into two categories based on the type of textual query (Fig.~\ref{fig:12} and ~\ref{fig:14}):
\begin{enumerate}
	\item Discrete attribute-based retrieval.
	\item Natural language description based retrieval.
\end{enumerate}

A discrete attribute-based person retrieval system searches for the person using soft biometrics attributes, e.g., clothing colour, clothing type, and gender. While a natural language description-based system accepts a natural description, e.g., \textit{A woman is with average height wearing a white midi with no sleeve. She has black open hair}.

Applications like obvious question answering, image captioning, and person retrieval are gaining much attention due to the association of two widely researched domains, i.e., computer vision and natural language processing. Such applications expect the learning of discriminative feature representations from both images and text. The person retrieval algorithm ranks gallery images or person(s) in the surveillance frame according to their relevance to the description. The best matching images from the gallery or surveillance frame return the person of interest. This class of surveillance problems is known as text-based person search or person retrieval using natural language descriptions \cite{R85,R139}. Such a person retrieval problem aims to enable retrieval of the person of interest using two different domain modalities, i.e., text and image. It takes one type of data as an input query (i.e., text). It retrieves the relevant data of another type (i.e., person image) by mapping textual and visual features (Sec.~\ref{sec:2.3}). Such kind of retrieval is also known as \textit{cross-modal retrieval}  \cite{R139,R140,R141,R142}.

Among all person retrieval methodologies, image gallery-based techniques are not preferable for end-to-end system-level evaluations. They do not consider challenges like occlusion, pose, and illumination in person detection from the full surveillance frame. Natural language description-based person retrieval systems are more suited for real-time person retrieval from surveillance videos. A person retrieved using a textual query is an input query to image-based person ReID systems for re-identification of the same target in the camera network. This paper discusses soft biometric attribute-based person retrieval methodologies in a further section.

\subsection{Discrete attribute-based person retrieval}
\label{sec:3.1}

Soft biometric attributes like clothing colour and clothing type do not stay consistent for a given individual for a length of time. Such soft attributes keep on changing over a short period. Thus, soft biometric attribute-based person retrieval methods are \textit{short-term} retrieval methods \cite{R71,R119}. Such methods are well suited for applications like criminal investigation and searching for missing persons for a limited period. Discrete attribute-based person retrieval methods are further divisible into two categories based on how the features are extracted, i.e.,

\begin{enumerate}
	\item Handcrafted feature-based person retrieval.
	\item Deep feature-based person retrieval.
\end{enumerate}

\subsubsection{Handcrafted feature-based methodologies}
\label{sec:3.1.1}

\begin{figure*}
	\centering
	\includegraphics[width=\textwidth]{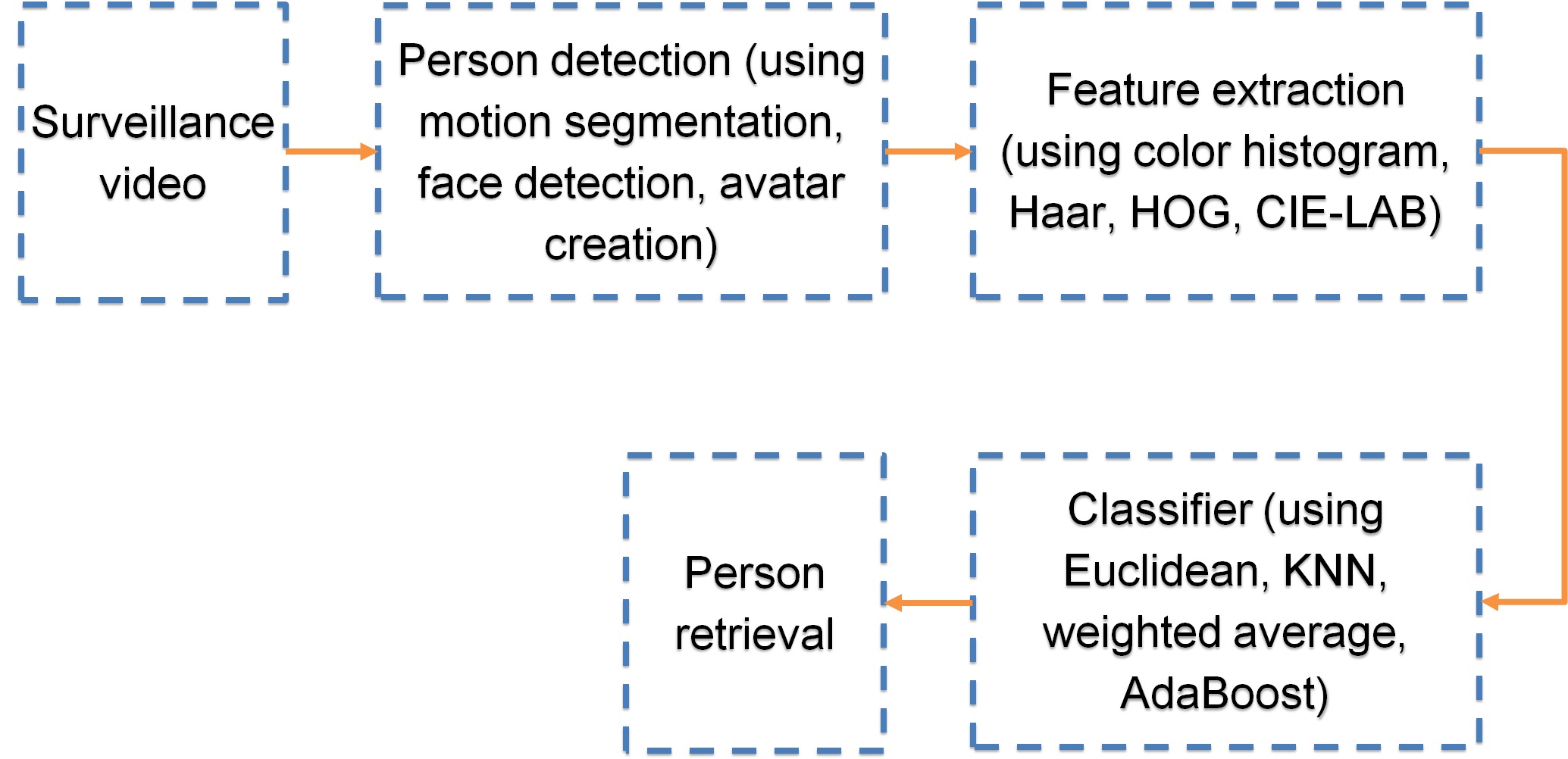}
	\caption{Block diagram for handcrafted feature-based methods.}
	\label{fig:15}       
\end{figure*}

Research before the era of deep learning shows promising methods for person retrieval based on handcrafted features. Fig.~\ref{fig:15} shows the general block diagram for person retrieval methods that use handcrafted features. Person detection is done using adaptive background segmentation, face detection, frame differencing, and query-based avatar creation. Feature extraction is a critical step where hand-engineered features are extracted using popular algorithms. Further, the feature fusion and classification are done for target person retrieval.

Vaquero et al. \cite{R71} exploit the limitation of the sensitivity of face recognition technology against illumination changes, low-resolution videos, and pose variations. They are the first to implement a video-based visual surveillance system that uses a person’s fine-grained parts and attributes. Person detection uses a face detector. Further, the body is divisible into three regions with soft biometrics from the areas, i.e., face (hair type, eyewear type, facial hair type), torso (clothing color), and leg (clothing color). The nine facial attributes are extracted by training an individual Viola-Jones detector using Haar features. A normalized colour histogram is in hue, saturation and luminance (HSL) space for each body part (i.e., torso and leg).

Denman et al. \cite{R73} also propose a body part segmentation-based approach that detects a person using adaptive background segmentation \cite{R91} and segments the body into three parts (head, torso, and leg) using an average gradient across rows of the image. Calculation of height uses the Tsai \cite{R120} camera calibration approach and colour histograms for clothing colour. These features are fused using a weighted sum fusion approach. The algorithm evaluation on the PETS \cite{R121} dataset achieves the best results with an Equal Error Rate (EER) of  26.7\% (colour model), the worst EER rate of 39.6\% (size model), and an EER rate of 29.8\% (a combination of colour and size).

The methods in \cite{R71,R73} detect a person first and then match the query rather than searching the person based on the query. Such limitation is removable by creating an avatar of the person from discrete soft biometric attributes  \cite{R74}. The particle filter is then applied to drive a search in the surveillance frame. The methods mentioned here propose a new dataset for soft biometric-based person localisation. The height of a person is broken up into 5 classes (concise, short, average, tall, and very tall) and colour into 11 classes (black, blue, brown, green, grey, orange, pink, purple, red, yellow, and white). Localisation accuracy is measurable using Intersection-over-Union (IoU). These methods achieve 24.28\% average localisation accuracy.

Pedestrian semantic attribute-based ApiS dataset is in \cite{R110} with 11 binary attributes and two multiclass attributes. AdaBoost classifiers are useful for binary attribute classification, and K Nearest Neighbours (KNN) classifier is useful for multiclass attribute classification. This paper focuses on attribute classification and is useful for person retrieval. Halstead et al. \cite{R75} further extend the surveillance dataset of \cite{R74} by annotating 12 multiclass attributes on 110 persons. The authors improve the performance of \cite{R74} by adding more attributes for avatar creation and considering the  unequal reliability of each attribute. A 21\% improvement is reported over the baseline \cite{R74}. Avatar further extends in the form of channel representation (CR) \cite{R76}. The algorithm also incorporates shape information in the form of HOG representation to improve the performance. CR captures the spatial characteristics of colour and texture by representing them in the form of a multi-dimensional image. Each channel in CR represents either colour or texture information. The CR based approach achieves 44\% average localisation accuracy.

Martinho et al. \cite{R106} generate a feature vector of length 4704 for each 256 $\times$ 256 pre-processed image. They use CIE-LAB, and the Gabor channel filters feature for the generation of an Ensemble of Localised Features (ELF) descriptor. A HOG based descriptor is also useful for the creation of 2304 features. The Extra-Trees (ET) supervised ensemble learning algorithm is applied for classification. The algorithm achieves 12.5\% and 20.1\% at rank-1 accuracy for one-shot and multi-shot identification. Shah et al. \cite{R70} use clothing colour and clothing type for person retrieval. They detect the person using motion segmentation based on frame differencing. Background clutter is removable by a morphological operation. The ISCC-NBS colour model \cite{R122} and CIEDE2000 \cite{R123} distance metrics are useful for colour classification. Their method achieves a 67.96\% TP rate, which is better than the GMM (44.88\%) approach. Table~\ref{tab:3} shows an overview of the handcrafted feature-based methodologies. The performance column shows the highest value reported in the relevant literature in multiple scenario-based analyses.

\begin{landscape}
	\begin{table*}
		\caption{Overview of soft biometric attribute based person retrieval methodologies based on handcrafted features.}
		\label{tab:3}
		\begin{tabular}{p{0.12\columnwidth}p{0.08\columnwidth}p{0.08\columnwidth}p{0.12\columnwidth}p{0.12\columnwidth}p{0.12\columnwidth}p{0.12\columnwidth}}
			\hline\noalign{\smallskip}
			\textbf{Work} & \textbf{Year} & \textbf{\# Soft attributes} & \textbf{Features} & \textbf{Classifier} & \textbf{Dataset} & \textbf{Performance} \\
			
			\noalign{\smallskip}\hline\noalign{\smallskip}
			
			Vaquero et al. \cite{R71} & 2009 & 5 & Haar, normalized color histogram & Euclidean & Private & \textendash \\
			
			Denman et al.\cite{R73} & 2009 & 3 & Height, color histogram & Weighted sum & PETS \cite{R121} & EER (26.7\%) \\
			
			Denman et al. \cite{R74} & 2012 & 3 & Height, CIE-LAB & Weighted average & Private & IoU (24.28\%) \\
			
			Zhu et al. \cite{R110} & 2013 & 13 & MB-LBP histogram, HOG & AdaBoost, KNN & APiS \cite{R110} & \textendash \\
			
			Halstead et al.\cite{R75} & 2014 & 12 & Height, CIE-LAB & Weighted average & SAIVT \cite{R75} & IoU (29.38\%) \\
			
			Denman et al.\cite{R76} & 2015 & 12 & HOG, Color, shape, height & Weighted average & SAIVT\cite{R75} & IoU (30.00\%) \\
			
			Martinho et al.\cite{R106} & 2016 & 12 & CIE-LAB, HOG, ELF & ETC & SoBiR \cite{R106} & nAUC (88.1\%) \\
			
			Shah et al.\cite{R70} & 2017 & 3 & Color histogram & CIEDE2000 & Private & TP (67.96\%) \\

			\noalign{\smallskip}\hline\noalign{\smallskip}
			\noalign{\smallskip}
			
			\multicolumn{7}{p{0.76\columnwidth}}{\textbf{Abbreviations used:}} \\
			
			\noalign{\smallskip}
			\noalign{\smallskip}
			
			\multicolumn{3}{p{0.28\columnwidth}}{EER = Equal Error Rate} & \multicolumn{2}{p{0.24\columnwidth}}{IoU = Intersection-over-Union} & \multicolumn{2}{p{0.24\columnwidth}}{HoG = Histograms of Oriented Gradients} \\
			
			\multicolumn{3}{p{0.28\columnwidth}}{MB-LBP = multi-scale block local binary patterns} & \multicolumn{2}{p{0.24\columnwidth}}{KNN = K - Nearest Neighbors} & \multicolumn{2}{p{0.24\columnwidth}}{TP = True Positive rate} \\
			
			\multicolumn{3}{p{0.28\columnwidth}}{ELF = Ensemble of Localised Features} & \multicolumn{2}{p{0.24\columnwidth}}{ETC = Extra Tree Classification} & \multicolumn{2}{p{0.24\columnwidth}}{nAUC = normalised Area Under Curve} \\
			
			\noalign{\smallskip}\hline\noalign{\smallskip}
			
		\end{tabular}
	\end{table*}
\end{landscape}

With the advent of deep learning techniques, person retrieval methods have seen a significant improvement in accuracy measures. Most state-of-the-art person retrieval methods are based on deep learning architectures.

\subsubsection{Deep feature-based methodologies}
\label{sec:3.1.2}

Deep learning-based methodologies are becoming popular in the past few years due to their efficient feature learning ability. The deep Convolutional Neural Network (DCNN) based approach has gained more attention in the computer vision community.  Table~\ref{tab:4} shows an overview of the deep feature-based person retrieval methodologies. The performance column shows the highest value reported in the relevant literature in the case of multiple scenario-based analyses. Semantic Retrieval Convolutional Neural Network (SRCNN) developed by Martinho et al. in \cite{R124} shows the evaluation of a similar setup of \cite{R106}. Binary Cross-Entropy (BCE) and Mean Squared Error (MSE) loss functions quantify binary classification and regression. SRCNN achieves 35.7\% and 46.4\% at rank-1 accuracy for one-shot and multi-shot identification, respectively. Thus, a deep feature based SRCNN approach demonstrates a rank-1 accuracy improvement of 23.2\% and 26.3\% over a handcrafted feature-based system of \cite{R106}.

The baseline method \cite{R76} of AVSS 2018 challenge II \cite{R5} dataset is implementable using handcrafted features, while all participants \cite{R69,R79,R80} of the challenge have evaluations based on deep features. The following discussions in this section consist of methodologies implemented on AVSS 2018 challenge II \cite{R5} dataset as well as some other approaches on large scale datasets like Market-1501 \cite{R105}, DukeMTMC \cite{R107}, PA100K \cite{R117} and CUHK03 \cite{R113}.

Galiyawala et al. \cite{R69} use height, clothing colour, and gender for person retrieval. Person detection and semantic segmentation using Mask R-CNN \cite{R98} help to remove the cluttered background. It results in a clutter-free torso patch for efficient colour classification. Height is available using a camera calibration approach \cite{R120}. Torso clothing colour and gender are classified using a fine-tuned AlexNet \cite{R125} based individual model. Sequential implementation of height, colour, and gender filter aims to eliminate the detected persons and leave only the target person. This linear filtering-based approach achieves an average IoU of 0.36. Schumann et al. \cite{R80} detect the person using the Single-Shot Multibox Detector (SSD) \cite{R126}. Early-stage false positives are eliminated by background modelling based on a mixture of Gaussian distribution. The strategy of the ensemble of classifiers is adapted, and predictions are fused by computing mean or weighted mean. The evaluation is based on the Euclidean distance between a query vector and each detected person's attribute probability to produce the final result.

\begin{landscape}
	\begin{table*}
		\caption{Overview of soft biometric attribute based person retrieval methodologies based on deep features.}
		\label{tab:4}
		\begin{tabular}{p{0.12\columnwidth}p{0.08\columnwidth}p{0.1\columnwidth}p{0.2\columnwidth}p{0.16\columnwidth}p{0.12\columnwidth}}
			
			\hline\noalign{\smallskip}
			
			\textbf{Work} & \textbf{Year} & \textbf{\# Soft attributes} & \textbf{Deep network} & \textbf{Dataset} & \textbf{Performance} \\
			
			\noalign{\smallskip}\hline\noalign{\smallskip}
			
			Martinho et al.\cite{R124} & 2016 & 12 & SRCNN \cite{R124} & SoBiR \cite{R106} & nAUC (92.8\%) \\
			
			Galiyawala et al.\cite{R69} & 2018 & 3 & Mask R-CNN \cite{R98}, AlexNet \cite{R125} & AVSS 2018 Challenge II \cite{R5} & Avg. IoU (0.363) \\
			
			Schumann et al.\cite{R80} & 2018 & 9 & SSD \cite{R126}, MobileNet-v1 \cite{R128}, ResNet-50-v1 \cite{R127}, DenseNet-121 \cite{R129}, DenseNet-169 \cite{R129} & AVSS 2018 Challenge II \cite{R5} & Avg. IoU (0.503) \\
			
			Yaguchi et al.\cite{R79} & 2018 & 9 & Mask R-CNN \cite{R98}, DenseNet-161 \cite{R129} & AVSS 2018 Challenge II \cite{R5} & Avg. IoU (0.511) \\
			
			Galiyawala et al.\cite{R78} & 2019 & 4 & Mask R-CNN \cite{R98}, DenseNet-169 \cite{R129} & AVSS 2018 Challenge II \cite{R5} & Avg. IoU (0.569) \\
			
			Sun et al.\cite{R81} & 2018 & 30 & PCB \cite{R81}, ResNet-50 \cite{R127} & Market-1501 \cite{R105}, DukeMTMC \cite{R107}, CUHK03 \cite{R113} & Rank-1 acc. (92.3\%) \\
			
			Dong et al.\cite{R83} & 2019 & 30 & AIHM \cite{R83}, ResNet-50 \cite{R127} & Market-1501 \cite{R105}, DukeMTMC \cite{R107}, PA100K \cite{R117} & Rank-1 mAP. (50.5\%) \\

			\noalign{\smallskip}\hline\noalign{\smallskip}
			\noalign{\smallskip}
			
			\multicolumn{6}{p{0.78\columnwidth}}{\textbf{Abbreviations used:}} \\
			
			\noalign{\smallskip}
			\noalign{\smallskip}
			
			\multicolumn{2}{p{0.20\columnwidth}}{IoU = Intersection-over-Union} & \multicolumn{2}{p{0.3\columnwidth}}{SRCNN = Semantic Retrieval Convolutional Neural Network} & \multicolumn{2}{p{0.28\columnwidth}}{SSD = Single-Shot Multibox Detector} \\
			
			\multicolumn{2}{p{0.20\columnwidth}}{TP = True Positive rate} & \multicolumn{2}{p{0.3\columnwidth}}{AIHM = Attribute-Image Hierarchical Matching} & \multicolumn{2}{p{0.28\columnwidth}}{PCB = Part-based Convolutional Baseline} \\
			
			\multicolumn{2}{p{0.20\columnwidth}}{nAUC = normalised Area Under Curve} & \multicolumn{2}{p{0.3\columnwidth}}{DenseNet = Densely connected convolutional networks} & \multicolumn{2}{p{0.28\columnwidth}}{ResNet = Residual Neural Network} \\
			
			\noalign{\smallskip}\hline\noalign{\smallskip}
			
		\end{tabular}
	\end{table*}
\end{landscape}

Yaguchi et al. \cite{R79} also use mask R-CNN for person detection and DenseNet-161 for attribute classification. Initially, they estimate all the attributes of the detected persons, and then the matching score is calculated using a Hamming loss. The person with the minimum loss is the target. Galiyawala et al. \cite{R78} further improve the linear filtering approach of \cite{R69} by introducing adaptive torso patch extraction and bounding box regression. Torso patch extraction is undertaken by deciding the torso region according to clothing type attribute. Thus, it removes noisy pixels from the torso patch and provides better colour classification. An IoU based box regression predicts the bounding box in the frame where soft biometric attribute-based retrieval fails. The approaches in \cite{R69} and \cite{R78} follow a linear filtering approach that filters out the non-matching person according to attributes and leaves the target in the end. The other two methods in \cite{R79} and \cite{R80} estimate all the detected attributes of a person in parallel. These methods fuse the characteristics in the end to retrieve the target. The linear filtering approach does not need to estimate all the attributes for all detection persons. However, an error in the first filter will propagate to further attribute filtering and reduce retrieval accuracy.

\begin{figure*}
	\centering
	\includegraphics[width=\textwidth]{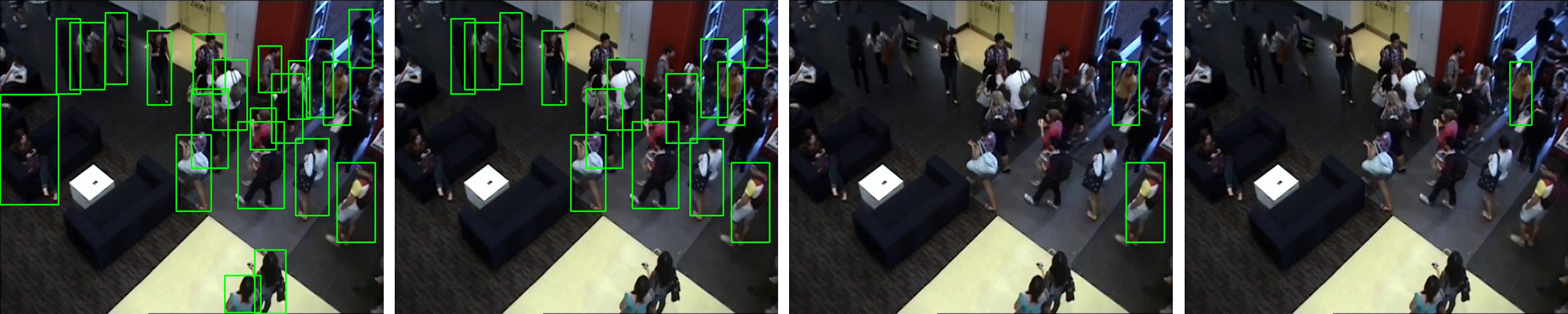}
	\caption{Sample output frames for the person retrieved using semantic description by approach in \cite{R78}: person from Test Sequence 11 and Frame 66 with semantic description height (very short; 140-160 cm), torso type (short sleeve), torso colour-1 (yellow), torso colour-2 (NA) and gender (male). Images from left to right are Mask R-CNN person detection, height filtering, clothing colour filtering and gender filtering.}
	\label{fig:16}       
\end{figure*}

The AVSS 2018 challenge II \cite{R5} dataset is evaluated based on two metrics; an average IoU and percent of frames with an IoU $\geq$ 0.4. State-of-the-art average IoU of 0.569 is achieved by a linear filtering approach \cite{R78}. 75.9\% of frames with an IoU $\geq$ 0.4 is achieved by \cite{R80}. Some sample qualitative results of person retrieval using the method in \cite{R78} are shown in Fig.~\ref{fig:16}.  It offers a surveillance frame from AVSS 2018 challenge II datasets for test sequence 40 and frame 66 with a semantic description, namely, height (very short; 140-160 cm), torso type (short sleeve), torso colour-1 (yellow), torso colour-2 (NA) and gender (male). The first image shows the person detection output using Mask R-CNN. Height filter output shows that many persons match the given query. Among these persons, the colour filter output produces a couple of matches and this is further refined by using a gender filter. It is to be noted that all deep features-based approaches \cite{R59,R78,R79,R80} perform better than the handcrafted feature-based baseline approach \cite{R76}.

Sun et al. \cite{R81} apply part level features because they provide fine-grained information for person description. The authors propose a Part-based Convolutional Baseline (PCB) network for part-based feature extraction. Part-level features are learnable by conducting uniform partition on the conv-layer without explicitly partitioning images. The spatial consistency within-part is exploitable to refine the coarse partition provided by PCB. The improvement in the uniform partition is made by the Refined Part Pooling (RPP) network. It achieves 92.3\% of rank-1 accuracy with only the PCB. Further, it improves to 93.8\% of rank-1 accuracy by employing RPP with PCB network on the Market-1501 \cite{R105} dataset. The text attribute combinations exist on a large scale in a real scenario. However, a minimal amount of combination with sufficient data is available for training. Except for such combinations, others are never modelled during training. Thus, Dong et al. \cite{R83} formulate a textual attribute query-based person search problem as a zero-shot (ZSL) learning problem for the first time. The authors propose Attribute-Image Hierarchical Matching (AIHM), which matches attributes and images at multiple hierarchical levels. The algorithm achieves state-of-the-art rank-1 mAP of 43.3\% on Market-1501  \cite{R105}, 50.5\% on DukeMTMC \cite{R107} and 31.3\% on PA100K \cite{R117} datasets. Table~\ref{tab:5} shows a comparison of handcrafted and deep feature-based methods with their advantages and disadvantages.

\begin{table*}
	\begin{center}
		\caption{Comparison of handcrafted and deep feature based methods.}
		\label{tab:5}       
			\begin{tabular}{p{0.12\columnwidth}p{0.12\columnwidth}p{0.66\columnwidth}}
				\hline\noalign{\smallskip}
				\textbf{Category} & \textbf{Sub class} & \textbf{Characteristics Advantages/disadvantages} \\
				\noalign{\smallskip}\hline\noalign{\smallskip}
				\textbf{Hand crafted features} & Haar \cite{R71} & Face detection works well for near view but it is difficult to detect for far view and crowded scenarios. \\
				
				& Color histogram \cite{R70,R71,R73} & Simple to implement and supports the less number of colors (i.e. primary colors) for classification but difficult if color classes increases (different shades of color) as it involves quantization process for histogram calculations.\\
				
				& HOG \cite{R76,R110} & HOG is good descriptor but its coarseness leads to sub-region processing in image which increases computational complexity.\\
				
				& Height \cite{R73,R74,R75} & It is calculated based on camera calibration parameters, but a small error in parameters estimation leads to high error in height calculations. Moreover, the camera parameter estimation need to done again if camera is moved.\\
				
				\noalign{\smallskip}\hline
				
				\textbf{Deep features} & AlexNet \cite{R69} & It is small network and 61M parameters to learn, but the classification accuracy is low.\\
				
				& MobileNet [80] & It’s a light weight network and 4.2M parameters to learn, but it is having little high error rate.\\
				
				& ResNet-50 [80, 81, 83] & It is having advantage of residual block which is having skip connection from input for better learning ability. Over 25M parameters to learn.\\
				
				& DenseNet [78, 79, 80] & It is having connection from each layer to every other layer and hence leveraging information from other layers which learns much better compare to other networks.\\
				
				\noalign{\smallskip}\hline
			\end{tabular}
		\end{center}
	\end{table*}

A query-based personal retrieval system usually provides soft biometric keywords as the input query. Improvisation in natural language processing now allows automatic extraction of the keywords from sentences. It will move the system towards full automation.

\subsection{Natural language description-based person retrieval}
\label{sec:3.2}

Cross-modal retrieval-based applications are drawing attention due to the rapid growth of multimodal data like text, image, video, and audio. Features from different modalities like text and image are not directly comparable as they lie entirely in other spaces. Hence, it is a challenging problem due to the sizeable heterogeneous gap between different text and image modalities. One such issue of person retrieval from surveillance video using natural language description is in this section. Table~\ref{tab:6} shows an overview of different methodologies for person retrieval using natural language descriptions.

Zhou et al. \cite{R89} develop an attention-based algorithm that localises a person in the surveillance frame using attributes and natural language query. The author annotated the cityscapes dataset \cite{R132} surveillance frame with attributes and descriptions because the dataset did not have natural language descriptions. Matrix representations of sentence expressions use the Skip-gram model \cite{R143}. Attributes and descriptions are by bidirectional Long-Short Term Memory (BLSTM) \cite{R130,R131} network. Visual features extraction is done using Faster R-CNN \cite{R97} and ResNet152 \cite{R127} with the algorithm achieving 74.6\% recall@1 i.e., 74.6\% of the highest scoring box is correct. The Cityscape dataset contains only street views, i.e., frontal view camera. Hence, it does not cover various view challenges for surveillance. Also, the description annotated dataset is not available publicly and it limits the usability.

Li et al. \cite{R85} first published a large-scale person dataset with natural language description, CUHK-PEDES (publicly available), as discussed in Sect.~\ref{sec:2.4}. The author proposes a Recurrent Neural Network with a Gated Neural Attention mechanism (GNA-RNN) to establish the baseline on CUHK-PEDES. The network consists of a visual sub-network for visual feature extraction from an image and a language sub-network for textual feature extraction from a description. The visual sub-network consists of VGG-16 \cite{R133} as a backbone network and generates 512 graphical units. Each optical unit determines the existence of a specific appearance pattern. RNN with Long Short-Term Memory (LSTM) is useful in language sub-network, which takes words and images as input. It outputs unit level attention that decides which visual units should pay more attention to the word. The word-level gate determines the importance of the word, e.g., the word \textquotedblleft white\textquotedblright has more weightage than the word \textquotedblleft the\textquotedblright . The GNA-RNN network is trained in an end-to-end manner and provides a top-1 accuracy of 19.05\%. This approach creates the baseline for further research in cross-modal person retrieval based on natural language descriptions.

\begin{landscape}
	\begin{table*}
		\caption{Overview of natural language description based person retrieval methodologies.}
		\label{tab:6}
		\begin{tabular}{p{0.14\columnwidth}p{0.08\columnwidth}p{0.14\columnwidth}p{0.14\columnwidth}p{0.12\columnwidth}p{0.12\columnwidth}p{0.12\columnwidth}}
			\hline\noalign{\smallskip}
			
			\textbf{Work} & \textbf{Year} & \multicolumn{2}{p{0.28\columnwidth}}{\hspace{1.3cm}\textbf{Deep network}} & \textbf{Method} & \textbf{Dataset} & \textbf{Performance} \\
			
			\noalign{\smallskip}\hline\noalign{\smallskip}
			
			& & \textbf{Visual features from person image} & \textbf{Text features from natural language description} & & & \\
			
			\noalign{\smallskip}\hline\noalign{\smallskip}
			
			Zhou et al.\cite{R89} & 2017 & Faster R-CNN \cite{R97}, Resnet152 \cite{R127} & Bi-LSTM \cite{R130,R131} & \textendash & Cityscapes \cite{R132}, Private & Recall@1 (74.6\%) \\ 
			
			Li et al.\cite{R85} & 2017 & VGG-16 \cite{R133} & LSTM \cite{R130} & GNA-RNN \cite{R85} & CUHK-PEDES \cite{R85} & Top-1 acc. (19.05\%) \\
			
			Li et al.\cite{R84} & 2017 & VGG-16 \cite{R133} & LSTM \cite{R130}, word2vec [144] & IATV [84] & CUHK-PEDES \cite{R85} & Top-1 acc. (25.94\%) \\
			
			Chen et al.\cite{R82} & 2018 & ResNet50 \cite{R127} & LSTM \cite{R130} & GLIA \cite{R82} & CUHK-PEDES \cite{R85} & Top-1 acc. (43.58\%) \\
			
			Zheng et al.\cite{R134} & 2017 & ResNet-50 \cite{R127} & Text-CNN \cite{R134}, word2vec \cite{R144} & DPCE \cite{R134} & CUHK-PEDES \cite{R85} & Top-1 acc. (44.40\%) \\
			
			Niu et al.\cite{R135} & 2019 & VGG-16 \cite{R133}, ResNet-50 \cite{R127} & Bi-GRU \cite{R145,R146} & MIA \cite{R135} & CUHK-PEDES \cite{R85} & Recall@1 (48.00\%) \\
			
			Zhang et al.\cite{R136} & 2018 & MobileNet \cite{R128} & Bi-LSTM \cite{R130,R131} & CMPC + CMPM \cite{R136} & CUHK-PEDES \cite{R85} & Recall@1 (49.37\%) \\
			
			Wang et al.\cite{R137} & 2019 & MobileNet \cite{R128} & Bi-LSTM \cite{R130,R131} & MCCL \cite{R137} & CUHK-PEDES \cite{R85} & Top-1 acc. (50.58\%) \\
			
			Sarafianos et al.\cite{R138} & 2019 & ResNet-101 \cite{R127} & BERT \cite{R148}, LSTM \cite{R130} & TIMAM \cite{R138} & CUHK-PEDES \cite{R85} & Top-1 acc. (54.51\%) \\
			
			Aggarwal et al.\cite{R139} & 2020 & MobileNet \cite{R128} & NLTK \cite{R149}, Bi-LSTM \cite{R130,R131} & CMAAM \cite{R139} & CUHK-PEDES \cite{R85} & Top-1 acc. (56.68\%) \\
			
			\noalign{\smallskip}\hline\noalign{\smallskip}
			\noalign{\smallskip}
			
			\multicolumn{7}{p{0.88\columnwidth}}{\textbf{Abbreviations used:}} \\
			
			\noalign{\smallskip}
			\noalign{\smallskip}
			
			\multicolumn{3}{p{0.36\columnwidth}}{BERT = Bidirectional Encoder Representations from Transformers} & \multicolumn{2}{p{0.26\columnwidth}}{CUHK-PEDES = CUHK Person Description Dataset} & \multicolumn{2}{p{0.24\columnwidth}}{MCCL = Mutually Connected Classification Loss} \\
			
			\multicolumn{3}{p{0.36\columnwidth}}{Bi-GRU = Bi-directional gated recurrent unit} & \multicolumn{2}{p{0.26\columnwidth}}{DPCE = Dual-Path Convolutional Image-Text Embeddings} & \multicolumn{2}{p{0.24\columnwidth}}{MIA = Multi-granularity Image-text Alignments} \\
			
			\multicolumn{3}{p{0.36\columnwidth}}{Bi-LSTM = Bidirectional Long-Short Term Memory} & \multicolumn{2}{p{0.26\columnwidth}}{GLIA = Global and Local Image-language Association} & \multicolumn{2}{p{0.24\columnwidth}}{GNA-RNN = Recurrent Neural Network with Gated Neural Attention} \\
			
			\multicolumn{3}{p{0.36\columnwidth}}{CMAAM = Cross Modal Attribute Aided Matching} & \multicolumn{2}{p{0.26\columnwidth}}{IATV = Identity-Aware Textual-Visual Matching} & \multicolumn{2}{p{0.24\columnwidth}}{TIMAM = Text-Image Modality Adversarial Matching} \\
			
			\multicolumn{3}{p{0.36\columnwidth}}{CMPM+CMPC = Cross-Modal Projection Matching loss and a Cross-Modal Projection Classification loss} & \multicolumn{2}{p{0.26\columnwidth}}{LSTM = Long-Short Term Memory} & \multicolumn{2}{p{0.24\columnwidth}}{ } \\
			
			\noalign{\smallskip}\hline\noalign{\smallskip}
			
		\end{tabular}
	\end{table*}
\end{landscape}

The approaches discussed below are evaluated on this baseline. Most of the large-scale dataset contains identity level annotations considered by Li et al. \cite{R84} to match visual and textual domains. Identity-aware textual-visual matching is done in a two-stage network. Identity-level annotations are effectively utilized by introducing a Cross-Modal Cross-Entropy (CMCE) loss in the stage-1 network. The CMCE loss implicitly maximizes inter-identity feature distances and minimizes intra-identity feature distances. However, the coupling between visual and textual features generated through CMCE loss is loose. Hence, the initial matching of stage-1 is further refined by stage-2 CNN-LSTM with an underlying co-attention mechanism that produces the final textual-visual matching confidence. This two-stage framework achieves 25.94\% of top-1 accuracy.

Global and local level image-language association (GLIA) is proposed by Chen et al. \cite{R82} to exploit the semantic information available in the description. The GLIA approach gains a significant boost to the top-1 accuracy from the baselines and achieves 43.58\% top-1 accuracy. Zheng et al. \cite{R134} focus on the limitation of ranking loss. The authors do not explicitly consider the feature distribution in a single modality. They overcome it by considering the problem as instance-level retrieval and propose the instance loss. Each image-text query pair is regarded as an instance and observed as a class during training to learn finer granularity. Instead of considering pre-trained models for feature extraction, the method view adopts end-to-end learning from the data itself. Dual-path CNN-CNN i.e. Dual-Path Convolutional Image-Text Embedding (DPCE) architecture is proposed instead of the CNN-RNN approach for image-text matching. DPCE \cite{R134} achieves 44.40\% of top-1 rank accuracy. 

Multi-granularity Image-text Alignments (MIA) framework \cite{R135} adopts a multiple granularities (i.e., global-global, global-local, and local-local alignments) based approach for better similarity evaluations between text and image. The global context of image and description matches global-global granularity. On the other hand, relations between the global context and local components establishes the global-local alignment. Visual human parts fit with noun phrases in the final local-local granularity. The algorithm achieves 48.00\% of recall@1. Zhang et al. \cite{R136} focus on learning discriminative features by proposing two loss functions, i.e., cross-modal projection matching (CMPM) loss and a cross-modal projection classification (CMPC) loss. Natural language description is first tokenized into words and processed sequentially using Bi-LSTM. Visual features are extractable from the last pooling layer of MobileNet \cite{R128}. The association module embeds visual and textual elements into a shared latent space. 49.37\% of recall@1 is achieved by the algorithm while considering both CMPM and CMPC losses. Similar to work in \cite{R84}, Wang et al. \cite{R137} also utilize identity-level information and propose Mutually Connected Classification Loss (MCCL). They first create a baseline approach before applying MCCL for better feature embedding. This baseline approach uses MobileNet pre-trained on ImageNet \cite{R147} for visual features, Bi-LSTM for textual element, and triplet loss function for cross-modal feature embedding. Triplet loss does not fully exploit feature distribution. The MCCL classification weight is shared between both modalities. Only the baseline approach with triplet loss achieves 45.55\% of recall@1 while MCCL achieves 50.58\% of recall@1.

The majority of approaches so far introduce a new loss function for the network to learn better feature representations without the complexity of textual phrases. For example, the word \textquotedblleft t-shirt\textquotedblright is an adjective, but it might be useful as a noun in the description. Such limitations go away by introducing a Text-Image Modality Adversarial Matching (TIMAM) framework  \cite{R138}. Sarafianos et al. \cite{R138} propose adversarial representation learning, which helps bring features from different modalities very close. TIMAM attains top-1 accuracy of 54.51\%. Aggarwal et al. \cite{R139} use attribute classification as an additional task and identity for bridging the gap between different modalities to improve representation learning. The method uses Deep Coral loss [150] to reduce the modality gap. They achieve state-of-the-art top-1 accuracy of 56.61\% on CUHK-PEDES dataset. Thus, most of the work \cite{R82,R84,R85,R134,R135,R136,R137,R138,R139} shows cross-modal person retrieval on the only publicly available CHUHK-PEDES dataset except \cite{R89}. CUHK-PEDES contains only the image gallery of cropped persons, limiting practical usage in real-time scenarios to retrieve the person from the input of full surveillance frames.

\section{Conclusions and open research challenges}
\label{sec:4}

Security of society has been a significant concern over the years, and biometric-based security has gained prominence due to exhaustive research in the field. Biometric security has evolved through the following stages:
\begin{itemize}
	\item Utilizing hard biometrics \cite{R15,R16}.
	\item Use of soft biometrics to improve the hard biometric-based systems \cite{R6}.
	\item Soft biometric attribute-based person retrieval \cite{R7,R8,R9,R10,R11,R12,R13}.
	\item Natural language description-based person retrieval \cite{R82,R84,R85,R134,R135,R136,R137,R138,R139} inherently uses soft biometrics in natural language description.
\end{itemize}

It has progressed from constrained environment-based fingerprint and face recognition systems to surveillance video-based person retrieval in the wild.

There are mainly two types of person retrieval methods seen in current research: discrete attribute-based (handcrafted feature-based and deep feature-based) methods and natural language description-based methods. Each has advantages and disadvantages. Handcrafted feature-based methods have limited performance due to less discriminative power of hand-engineered features and limited support to more classes for classification. Deep features are much more robust and produce better retrieval results but require a high amount of training data and high computation resources. Discrete attributes are easy to acquire, but do not suit real-time applications where the natural language descriptions are typically used. Most natural language description-based methods (Table~\ref{tab:6}) are not evaluated on full-frame surveillance videos. Moreover, the dataset is also a key element in any system development, including real-time scenarios. Hence, the current research hotspot is to develop an end-to-end person retrieval system from unconstrained surveillance videos.

Given the comprehensive review, some open problems that require further research and analysis are as follows:

\begin{enumerate}
	\item Diverse dataset development: This review focuses on person retrieval from surveillance videos using soft biometrics, which are in the form of either a discrete attribute or natural language description. Both forms of query are in the form of text to the person retrieval system. The development of various person retrieval methodologies would not have been possible without the continuous evolution of challenging datasets (Table~\ref{tab:2}) proposed by active researchers in the vision and language field. The datasets cover the following challenges: 
	\begin{itemize}
		\item Different resolutions
		\item Varying illumination conditions
		\item Occlusion
		\item Merging foreground with background
		\item Viewpoint variations
		\item Crowded scene
		\item Indoor and outdoor environments
		\item Time consistency/time duration
		\item Type of camera or multiple cameras
	\end{itemize}
	
	These challenges (discussed in Sec.~\ref{sec:2.1}) are not present in all datasets, but they lack one or the other difficulty. For example, CUHK-PEDES is the only large-scale dataset available for cross-modal person retrieval. However, it lacks full surveillance frames and samples of interest merging with background and crowd scenarios. 
	
	Issues involved in person detection in regular and crowded scenarios cannot be evaluated due to non-availability of full surveillance frames. AVSS 2018 challenge II is the only dataset that has low-resolution full surveillance frames and covers many challenges. However, it also lacks outdoor scenes and natural language descriptions. Thus, it also cannot be tested against various natural outdoor conditions and limits the development of a robust end-to-end cross-modal person retrieval framework. The more recent P-DESTRE dataset provides 4K videos but lacks in indoor environment. The majority of the datasets consist of very few samples with occlusion and a mixture of indoor and outdoor environment samples of the same person. Thus, this review of a promising large-scale dataset provides an opportunity to develop a more challenging dataset that can overcome the limitations of an individual dataset.
	
	\item Developing robust person attribute recognition model: As discussed in Sect.~\ref{sec:2.3}, person attribute recognition is the most critical task for visual attribute extraction. This task uses a cropped person dataset for model preparations. But the attribute model trained on one dataset may not achieve a good performance for other datasets due to lack of diversity. Differences in images from different datasets in terms of colour, view, pose, and illumination conditions are readily observable in Fig.~\ref{fig:17}. The model training has to be done by merging samples from a different dataset. But annotations to the images from other datasets are not the same. For example, RAP images are annotated with clothing type with 10 classes, namely, \{ub-Shirt, ub-Sweater, ub-Vest, ub-Tshirt, ub-Cotton, ub-Jacket, ub-SuitUp, ub-Tight, ub-ShortSleeve, ub-Other\}. In comparison, AVSS 2018 challenge II images are annotated with clothing type with 3 classes, namely, \{long sleeve, short sleeve, no sleeve\}. The solution is that these annotations need to be mapped by annotating one of the datasets with similar annotations to other dataset classes.
	
	\begin{figure*}
		\centering
		\includegraphics[width=0.5\textwidth,height=0.5\textwidth]{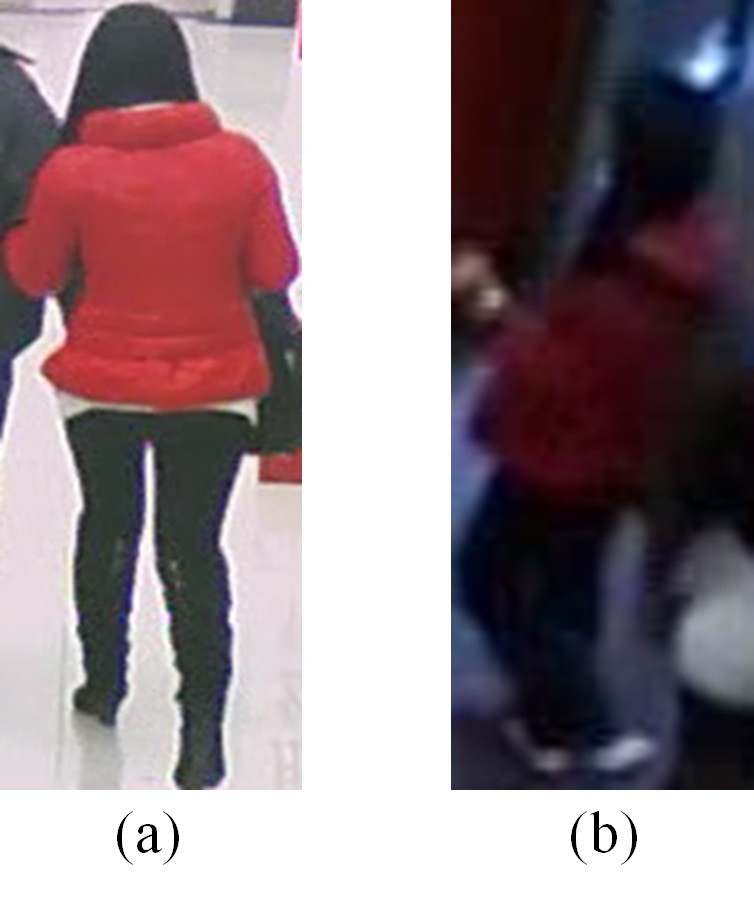}
		\caption{Images from two different dataset: (a) RAP \cite{R108}, (b) AVSS 2018 challenge II \cite{R5}.}
		\label{fig:17}       
	\end{figure*}
	
	Class imbalance is also a problem, e.g., samples for every clothing type may not be of the same number. The number of samples for clothing type in RAP dataset are \{14652, 5260, 3209, 15640, 7814, 19209, 2057, 2830, 6549, 1840\} \cite{R108}. Such a class imbalance problem can be resolved by weighting each class during the training. Assign a higher weight to a class with fewer samples and a lower weight to a class with more samples.
	
	The total number of training images in the RAP dataset is 67943, while AVSS has approximately 14000 images. It is observed from Fig.~\ref{fig:17} that the RAP dataset images have appropriate illumination conditions and resolutions. However, AVSS dataset images have challenging illumination conditions and lower resolutions. Thus, models trained by merging such datasets will produce good accuracy for clear images like RAP but not for images like the AVSS dataset. Such a problem can be resolved by generating images (for the dataset having fewer images) using domain adaption methodologies as discussed in \cite{R160,R161}.
	
	\item Robust framework development: Different methodologies are reviewed (Table~\ref{tab:3}, ~\ref{tab:4}, ~\ref{tab:5}) on the dataset mentioned in Table~\ref{tab:2}. State-of-the-art performance on AVSS 2018 challenge II datasets in terms of average IoU is 0.569 achieved by the linear filtering approach of \cite{R78} and an approach in \cite{R80} achieves 75.9\% of frames with an IoU $\geq$ 0.4. The method in \cite{R78} has a limitation of error propagation due to the sequential processing of samples and a significantly lower soft biometric amount. Such a process also require to create a separate model for individual attributes, e.g., gender model and colour model. It is not feasible and also a very costly solution to make a different model for every specific attribute. A single model for attribute recognition by multi-attribute learning (as discussed Sec.~\ref{sec:2.3}) helps resolve such a problem by considering the solutions suggested above to develop a robust person attribute recognition model.
	
	\item Adversarial learning-based approach: Although datasets are available with increasing complexity, e.g., number of persons, images, and attributes, such a cross-modal retrieval problem covers many possible real-scenarios, e.g., the backpack carried by a person in one camera view may not be visible in some other camera views. It is also expensive and tedious to cover many attribute combinations with image annotations in a dataset. Thus, image and text, both modalities together, produce many unseen scenarios (which are not available in the dataset) for person retrieval. It limits the usability and scalability of real-time deployments. Researchers in \cite{R156,R157,R158,R159} propose adversarial learning-based approaches to mitigate such issues. Yin et al. \cite{R156} suggest that attribute-based concept generation learns a discriminative joint space for image and text. Wu et al. \cite{R157}propose a deep adversarial data augmentation method with attribute (DADAA), which regularises training by generating unseen scenarios of person images. Unseen attribute combinations are used by a symbiotic adversarial learning (SAL) \cite{R159} framework to synthesize features of unseen categories and also to optimise feature embedding by cross-modal alignment in the common embedding space. Hence, data augmentation, unseen attribute combination and attribute-based image generation using adversarial learning for person retrieval are also exciting areas for further research.
	
	\item Backbone network explorations: The majority of the works reviewed in Tables~\ref{tab:4}, ~\ref{tab:5} use VGG-16, ResNet, DenseNet, and MobileNet, which are popular and provide state-of-the-art results. Nevertheless, the supremacy of CNN networks like EfficientNets \cite{R151} is as yet unexplored. None of the methods utilise EfficientNets, which are smaller and faster than VGG-16, ResNet, DenseNet, and MobileNet. For example, EfficientNet-B0 uses 5.3M parameters which are much less in comparison with ResNet-50 (26M) and DenseNet-169 (14M) while evaluating on ImageNet dataset \cite{R147}. Thus, a review of state-of-the-art methodologies and deep network creates a scope to develop a more robust network with better feature representation to improve performance. Hence, person retrieval in surveillance using a textual query is still a challenging and open research problem.
	
\end{enumerate}

In summary, soft biometrics bridges the gap between human language descriptions and machines to automate the person retrieval process. The application domain ranges from crime investigation, security to missing person retrieval. There are multiple soft biometric attributes that can be used for person retrieval, and thus the paper offers a way of selecting an attribute that will help the naive researcher begin research in the domain. The study includes the perspective of challenging datasets, leading edge methods, and deep networks. The paper presents an in-depth review by tracing the development of retrieval in chronological order using handcrafted features, simple attribute-based retrieval, and natural language-based retrieval. It guides beginners to gather essential ingredients for their research journey and, at the same time, provides an expert bird's eye view of the domain through a well-conceived summary. The goal of the comprehensive discussion undertaken in this paper about large-scale datasets and various existing problem definitions is to engender motivation to solve problems that exist in this domain.

%

\begin{acknowledgements}
The Board of Research in Nuclear Sciences (BRNS), Government of India (36(3)/14/20/2016-BRNS/36020) supports this work. The authors acknowledge the support of NVIDIA Corporation for a donation of the Quadro K5200 GPU used for this research. The authors are thankful to Ahmedabad University, India, for access to resources like GPUs. We would also like to thank the vision and language domain's active researchers for creating publicly available challenging datasets. 
\end{acknowledgements}

%
%




\end{document}